\documentclass[%
 reprint,
nofootinbib,
 amsmath,amssymb,
 aps,
 onecolumn,
]{revtex4-2}
\usepackage[utf8]{inputenc} 
\usepackage[T1]{fontenc}    
\usepackage{hyperref}       
\usepackage{url}            
\usepackage{booktabs}       
\usepackage{amsfonts}       
\usepackage{nicefrac}       
\usepackage{microtype}      
\usepackage{xcolor}         

\usepackage{amsmath}
\usepackage{amssymb}
\usepackage{graphicx}
\usepackage{hyperref}
\usepackage{mathtools}
\usepackage{url}
\usepackage{footnote}
\usepackage{comment}
\usepackage[figurename=Figure]{caption}

\usepackage{algorithm}
\usepackage{bbold}
\usepackage{algpseudocode}
\setcitestyle{numbers,square} 
\DeclareMathOperator{\E}{\mathbb{E}}
\DeclareMathOperator{\argmax}{\arg\max}

\usepackage{dcolumn}
\usepackage{bm}

\begin{document}

\preprint{APS/123-QED}

\title{Gibbs Sampling the Posterior of Neural Networks}

\author{Giovanni Piccioli}
\email{giovannipiccioli@gmail.com}
\author{Emanuele Troiani}%
\author{Lenka Zdeborová}
\affiliation{École Polytechnique Fédérale de Lausanne (EPFL)\\
  Statistical Physics of Computation Laboratory\\}

\begin{abstract}
  In this paper, we study sampling from a posterior derived from a neural network. 
  We propose a new probabilistic model consisting of adding noise at every pre- and post-activation in the network, arguing that the resulting posterior can be sampled using an efficient Gibbs sampler. 
  For small models, the Gibbs sampler attains similar performances as the state-of-the-art Markov chain Monte Carlo (MCMC) methods, such as the Hamiltonian Monte Carlo (HMC) or the Metropolis adjusted Langevin algorithm (MALA), both on real and synthetic data.
  By framing our analysis in the teacher-student setting, we introduce a thermalization criterion that allows us to detect when an algorithm, when run on data with synthetic labels, fails to sample from the posterior. 
  The criterion is based on the fact that in the teacher-student setting we can initialize an algorithm directly at equilibrium.
\end{abstract}
\maketitle
\section{Introduction}
\vspace{-3mm}
Neural networks are functions parametrized by the so-called weights, mapping inputs to outputs. Neural networks are commonly trained by seeking values of weights that minimize a prescribed loss function. In some contexts, however, we want to sample from an associated probability distribution of the weights. Such sampling is at the basis of Bayesian deep learning \cite{wilson2020bayesian,wang2020survey}. It is used in Bayesian uncertainty estimation \cite{kendall2017uncertainties,teye2018bayesian,maddox2019simple} or to evaluate Bayes-optimal performance in toy models where the data-generative process is postulated \cite{barbier2019optimal}.  In this paper, we focus on studying the algorithms and properties of such sampling. 

Given training inputs $X$ in Bayesian learning, one implicitly assumes the labels to be generated according to the stochastic process $y\sim P(y|X,W)$, where $W$ are the weight of the network, on which a prior $P(W|X)$ is placed. 
At its heart, Bayesian deep learning consists of sampling from the posterior probability of the parameters:
\begin{equation}
    \label{eq:bayes_theorem_nn}
    P(W|X,y)=\frac{P(y|W,X)P(W|X)}{P(y|X)},
\end{equation}
where we simply used Bayes theorem.
This sampling problem is, in general, NP-hard \cite{dagum1993approximating}, with many techniques being developed to sample from \eqref{eq:bayes_theorem_nn}. 
In this paper, we look at iterative algorithms that in the large time limit, return samples from the posterior distribution \eqref{eq:bayes_theorem_nn}. Most available algorithms for this task are based on MCMC methods. 
We focus on the two following questions:
\begin{itemize} 
    \item \textbf{Q1:} Do we have a method to evaluate whether the algorithms have thermalized, i.e., if the samples returned by the MCMC plausibly come from the posterior \eqref{eq:bayes_theorem_nn}?
    \item \textbf{Q2:} Which combinations of sampling algorithm and form of the posterior distribution achieve the best performance in terms of ability to thermalize while reaching a low test error?
\end{itemize}
The first question addresses the long-standing problem of estimating an MCMC's thermalization time, that is, the time at which the MCMC starts sampling well from the posterior. We propose a criterion for thermalization based on the teacher-student setting. The criterion can only be reliably applied to synthetic labels generated by a teacher network. 
After a comparison with other thermalization heuristics, we argue that the teacher-student criterion is more discriminative, in that it provides a higher lower bound to the thermalization time.
The second question explores the interplay between the form of the posterior and the sampling algorithm: since there is more than one way of translating a network architecture into a probabilistic process, we exploit this freedom to introduce a generative process in which noise is added at every pre- and post-activation of the network. We then design a Gibbs sampler tailored to this posterior and compare it to other commonly used MCMCs. 

\subsection{Related literature}
\vspace{-1mm}
\label{sec:related literature}

When running an MCMC one has to wait a certain number of iterations for the algorithm to start sampling from the desired probability measure. We will refer to this burn-in period as the thermalization time or $T_\text{therm} $\cite{newman1999monte}. Samples before $T_\text{therm}$ should therefore be discarded. 
Estimating $T_\text{therm}$ is thus of great practical importance, as it is crucial to know how long the MCMC should be run.

More formally, we initialize the MCMC at a starting state $W_0\in\mathcal W$ of our liking. We run the chain iteratively sampling $W(t)\sim P(W(t)|W(t-1))$, where $P(\cdot|\cdot)$ is the transition kernel.
If the kernel is ergodic and satisfies the relation $\pi(W)=\sum_{W'\in\mathcal W}P(W|W')\pi(W')$, for a certain probability measure $\pi(\cdot)$, then for $t\to\infty$ the MCMC will return samples from $\pi(\cdot)$. Thermalization is concerned with how soon the chain starts sampling approximately from $\pi(\cdot)$.

Consider an observable $\varphi:\mathcal W\mapsto \mathbb R$. Let $\pi_\varphi(\cdot)$ be the distribution of $\varphi(W)$ when $W\sim\pi(\cdot)$. 
Define $S_\varphi(\delta)\subseteq \mathbb R$ as the smallest set such that $\pi_\varphi(S_\varphi(\delta))\geq 1-\delta$. When $\delta\ll 1$, $S_\varphi(\delta)$ is a high probability set for $\varphi(\cdot)$. We can then look at thermalization from the point of view of $\varphi$. For a general initialization $W_0$ one will usually have $\varphi(W_0)\not\in S_\varphi(\delta)$, since in most initializations, $W_0$ is unlikely to be a typical sample form $\pi(\cdot)$. As more samples are drawn, the measure sampled by the chain will approach $\pi(\cdot)$, therefore we expect that $\varphi(W(t))\in S_\varphi(\delta)$ (up to a fraction $\delta$ of draws) for $t$ greater than some time $\hat t_\varphi=\hat t_\varphi(W_0)$. We call $\hat t_\varphi(W_0)$ the thermalization time of observable $\varphi$; notice that $\hat{t}_\varphi(W_0)$, depends both on the observable $\varphi$ and on the initial condition\footnote{In principle $\hat t_\varphi(W_0)$ also depends on the randomness of the MCMC. For the purpose of the discussion consider $\hat t_\varphi(W_0)$ to be an average over this randomness.}. In fact some observables may thermalize faster than others, and a good initialization can make the difference between an exponentially (in the dimension of $W$) long thermalization time and a zero one (for example if $W_0$ is drawn from $\pi(\cdot)$). In statistical physics it is common to say that the whole chain has thermalized when all observables that concentrate in the thermodynamic limit have thermalized \cite{newman1999monte,zdeborova2016statistical,mezard2009information}. This will be our definition of $T_\text{therm}$. 
Despite the theoretical appeal, this definition is inapplicable in practice. In fact, for most observables computing $\pi_\varphi(\cdot)$ is extremely hard computationally.

Practitioners have instead resorted to a number of heuristics, which provide lower bounds to the thermalization time. These heuristics usually revolve around two ideas. We first have methods involving multiple chains \cite{gelman1992inference, newman1999monte,brooks1998general,cowles1996markov}. In different flavours, all these criteria rely on comparing multiple chains with different initializations. Once all the chains have thermalized, samples from different chains should be indistinguishable. 
Another approach consists of finding functions with known mean under the posterior and verifying whether the empirical mean is also close to its predicted value \cite{gorham2015measuring,gorham2017measuring,fan2006output,cowles1996markov,zellner1995gibbs}. The proposed method for detecting thermalization relies instead on the teacher-student framework \cite{zdeborova2016statistical}. 

Another field we connect with is that of Bayesian learning of neural networks. For an introduction see \cite{jospin2022hands,goan2020bayesian,wang2020survey,magris2022bayesian} and references therein. We shall first examine the probabilistic models for Bayesian learning of neural networks  and then review the algorithms that are commonly used to sample. In order to specify the posterior \eqref{eq:bayes_theorem_nn}, one needs to pick the likelihood (or data generating process) $P(y|X,W)$. The most common model, employed in the great majority of works \cite{tishby1989consistent,izmailov2021bayesian,wilson2020bayesian,wenzel2020good,neal2012bayesian} is $P(y|X,W)=\frac{1}{Z}\exp\left(-\frac{1}{2\Delta}\sum_{\mu}\ell(y^\mu,f(X^\mu,W))\right)$, where $f(\cdot,W)$ is the neural network function, $\ell$ is the loss function, $\mu$ is the sample index, and $\Delta$ is a temperature parameter. As an alternative, other works have introduced the "stochastic feedforward networks", where noise is added at every layer's pre-activation \cite{neal1990learning,tang2013learning,yu2021simple,raiko2014techniques}. Outside of the Bayesian learning of neural networks  literature, models where intermediate pre- or post-activations are added as dynamical variables have also been considered in the predictive coding literature \cite{millidge2021predictive,millidge2022predictive,alonso2022theoretical,whittington2017approximation}.

Once a probabilistic model has been chosen, the goal is to obtain samples from the corresponding posterior. A first solution consists of approximating the posterior with a simpler distribution, which is easier to sample. This is the strategy followed by variational inference methods \cite{maddox2019simple,wang2016natural,tang2013learning,mackay1992practical,khan2018fast,wu2018deterministic}. 
Although variational inference yields fast algorithms, it is often based on uncontrolled approximations. Another category of approximate methods is that of "altered MCMCs", i.e., Monte Carlo algorithms which have been modified to be faster at the price of not sampling anymore from the posterior \cite{chen2014stochastic,nemeth2021stochastic,neal1992connectionist,ma2015complete,zhang2019cyclical,li2016preconditioned}. An example of these algorithms is the discretized Langevin dynamics \cite{welling2011bayesian}. Restricting the sampling to a subset of the parameters has also been considered in \cite{sharma2022bayesian} as an alternative training technique. 

Finally, we have exact sampling methods: these are iterative algorithms that in the large time limit are guaranteed to return samples from the posterior distribution. Algorithms for exact sampling mostly rely on MCMC methods. The most popular ones are HMC \cite{duane1987hybrid,neal2012bayesian}, MALA \cite{besag1994comments,roberts1996exponential,metropolis1953equation} and the No U-turn sampler (NUTS)\cite{hoffman2014no}. Within the field of Bayesian learning in neural networks, HMC is the most commonly used algorithm \cite{wenzel2020good,wilson2020bayesian,izmailov2021bayesian}. The proposed Gibbs sampler is inspired to the work of \cite{albert1993bayesian}, and later \cite{held2006bayesian,fruhwirth2010data}, that introduced the idea of augmenting the variable space in the context of logistic and multinomial regression.
\section{Teacher-student thermalization criterion}
\vspace{-3mm}
\label{sec:TS_sec} 
In this section we explain how to use the teacher-student setting to build a thermalization test for sampling algorithms. The test gives a lower bound on the thermalization time. We start by stating the main limitation of this approach: the criterion can only be applied to synthetic datasets. In other words, the training labels $y$ must be generated by a teacher network, using the following procedure.

We first pick arbitrarily the training inputs and organize them into an $n\times d$ matrix $X$. Each row $X^\mu$ of the matrix is a different training sample, for a total of $n$ samples. We then sample the teacher weights $W_\star$ from the prior $P(W)$. Finally, we generate the noisy training labels as $y^\mu\sim P(y|X^\mu,W_\star)$.
Our goal is to draw samples from the posterior $P(W|D)$, where  $D=\{(X^\mu,y^\mu)\}_{\mu\in[n]}$ indicates the training set.
Suppose we want to have a lower bound on the thermalization time of a MCMC initialized at a particular configuration $W_\text{start}$. The method consists of running two parallel chains $W_1(t)$ and $W_2(t)$. For the first chain, we use an informed initialization, meaning we initialize the chain on the teacher weights, thus setting $W_1(t=0)=W_\star$. For second chain we set  $W_2(t=0)=W_\text{start}$. To determine convergence we consider a test function $\varphi(W)$. We first run the informed initialization: after some time $T_1$, $\varphi(W_1(t))$ will become stationary. Using samples collected after $T_1$ we compute the expected value of $\varphi(\cdot)$ (let us call it $\overline{\varphi}$). Next, we run the second chain. The lower bound to the thermalization time of $W_2(t)$ is the time where $\varphi(W_2(t))$  becomes stationary and starts oscillating around $\overline{\varphi}$. In practice, this time is determined by visually inspecting the time series of $\varphi(\cdot)$ under the two initializations, and observing when the two merge.

At first glance this method does not seem too different from \cite{gelman1992inference} or \cite{brooks1998general}, whose method (described in Appendix \ref{app:Rhat}) relies on multiple chains with different initializations. There is however a crucial difference: under the informed initialization most observables are already thermalized at $t=0$. To see this, recall that the pair $W_\star, D$ was obtained by first sampling $W_\star$ from $P(W)$ and then sampling $D$ from $P(D|W_\star)$. This implies that $W_\star,D$ is a sample from the joint distribution $P(W,D)$. Writing $P(W|D)=\frac{P(W,D)}{P(D)}$, we see that $W_\star$ is also typical under the posterior distribution $P(W|D)$. In conclusion, the power of the teacher-student setting lies in the fact that it gives us access to one sample from the posterior, namely $W_\star$. It then becomes easier to check whether a second chain is sampling from the posterior by comparing the value of an observable. In contrast, other methods comparing chains with different initialization have no guarantee that if the two chains "merge" then the MCMC is sampling from the posterior, since it is possible that both chains are trapped together far from equilibrium.

\section{The intermediate noise model}
\vspace{-3mm}
\label{sec: intermediate_noise_model}
In this section, we introduce a new probabilistic model for Bayesian learning of neural networks. 
We start by reviewing the classical formulation of Bayesian learning of neural networks.
Let $f(x,W)$ be the neural network function, with $W$ its parameters, and $x\in\mathbb R^d$ the input vector. Given a training set $X\in\mathbb R^{n\times d},y\in\mathbb R^n$ we aim to sample from
\begin{align}
\label{eq:classical_posterior}
        & P(W|X,y)=\frac{1}{P(y|X)}P(W)\exp\left[-\frac{1}{2\Delta}\sum_{\mu=1}^n\ell\left(y^\mu,f(X^\mu,W)\right)\right],
\end{align}
where $\ell(\cdot,\cdot)$ is the single sample loss function, and $\Delta$ a temperature parameter. Notice that to derive \eqref{eq:classical_posterior} from \eqref{eq:bayes_theorem_nn}, we supposed that $P(W|X)=P(W)$, i.e., $W$ is independent of $X$. This is a common and widely adopted assumption in the Bayesian learning literature, and we shall make it in what follows.
Most works in the field of Bayesian learning of neural networks attempt to sample from \eqref{eq:classical_posterior}.
This form of the posterior corresponds to the implicit assumption that the labels were generated by
\begin{equation}
    \label{eq:classical_gen}
    y^\mu\sim P_\text{out}(y|f(X^\mu,W)),\text{ with } P_\text{out}(y|z)\propto e^{-\frac{1}{2\Delta}\ell(y,z)}
\end{equation} 
where $W$ are some weights sampled from the prior. 
We propose an alternative generative model based on the idea of introducing a small Gaussian noise at every pre- and post-activation in the network. The motivation behind this process lies in the fact that we are able to sample the resulting posterior efficiently using a Gibbs sampling scheme. Consider the case where $f(\cdot,W)$ is a multilayer perceptron with $L$ layers, without biases and with activation function $\sigma\left(\cdot\right)$. Hence we have
$f(x,W)= W^{(L)}\sigma\left(W^{(L-1)}\sigma\left(\dots \sigma(W^{(1)}x)\dots \right)\right)$. Here $W^{(\ell)}\in\mathbb R^{d_{\ell+1}\times d_\ell}$ indicates the weights of layer $\ell\in[L]$, with $d_\ell$ the width of the layer.  We define the pre-activations $Z^{(\ell)}\in\mathbb R^{n\times d_\ell}$ and post activations $X^{(\ell)}\in\mathbb R^{n\times d_\ell}$ of layer $\ell$. 
Using Bayes theorem and applying the chain rule to the likelihood we obtain
\begin{align}
    \label{eq:intermediate_noise_posterior}
    &\nonumber P(\{X^{(\ell)}\}_{\ell=2}^L,\{Z^{(\ell)}\}_{\ell=2}^L,\{W^{(\ell)}\}_{\ell=1}^L|X,y)=
    \\\nonumber&\frac{1}{P(y|X)}P(\{W^{(\ell)}\}_{\ell=1}^L)P(y,\{X^{(\ell)}\}_{\ell=2}^L,\{Z^{(\ell)}\}_{\ell=2}^L|\{W^{(\ell)}\}_{\ell=1}^L,X)
    =\frac{1}{P(y|X)}P(\{W^{(\ell)}\}_{\ell=1}^L)\times\\&\times \left[\prod_{\ell=2}^{L} P(Z^{(\ell+1)}|X^{(\ell)},W^{(\ell)})P(X^{(\ell)}|Z^{(\ell)})\right]P(Z^{(2)}|X,W^{(1)})
\end{align}
with the constraint $Z^{(L+1)}=y$. The conditional probabilities are assumed to be
\begin{align}
    \label{eq:posterior_flin}
    & P(Z^{(\ell+1)}|X^{(\ell)},W^{(\ell)})= \prod_{\mu=1}^n\prod_{\alpha=1}^{d_{\ell+1}}\mathcal{N}\left(Z^{(\ell+1)\mu}_\alpha\middle|W^{(\ell)T}_\alpha X^{(\ell)\mu},\Delta_Z^{(\ell+1)}\right)\\
    &P(X^{(\ell)}|Z^{(\ell)})=\prod_{\mu=1}^n\prod_{i=1}^{d_\ell}\mathcal N\left(X_{\alpha}^{(\ell)\mu}\middle|\sigma(Z^{(\ell)\mu}_i),\Delta_X^{(\ell)}\right),
\end{align}
where $\{\Delta^{(\ell)}_Z\}_{\ell=2}^{L+1},\,\{\Delta^{(\ell)}_X\}_{\ell=2}^{L}$ control the amount of noise added at each pre- and post- activation.
This structure of the posterior implicitly assumes that the pre- and post-activations are iteratively generated as 
\begin{equation}
    \label{eq:gen_process_intermediate_noise}
    Z^{(\ell+1)}= X^{(\ell)} W^{(\ell)T}+ \epsilon^{(\ell+1)}_Z, \quad X^{(\ell+1)}=\sigma(Z^{(\ell+1)})+ \epsilon^{(\ell+1)}_X, \quad \ell\in[L].
\end{equation}
$X^{(1)}=X \in \mathbb R^{n\times d}$ are the inputs, $Z^{(L+1)}=y\in \mathbb R^{n}$ represent the labels and $\epsilon_Z^{(\ell)}, \epsilon_X^{(\ell)}$ are $n\times d_{\ell}$ matrices of i.i.d. respectively $\mathcal{N}(0,\Delta^{(\ell)}_Z)$ and $\mathcal{N}(0,\Delta^{(\ell)}_X)$ elements.  We will refer to \eqref{eq:gen_process_intermediate_noise} as the intermediate noise generative process.

If we manage to sample from the posterior \eqref{eq:intermediate_noise_posterior}, which has been augmented with the variables $\{X^{(\ell)}\}_{\ell=2}^{L},\{Z^{(\ell)}\}_{\ell=2}^L$, then we can draw samples from $P(W|X,y)$, just by discarding the additional variables. A drawback of this posterior is that one has to keep in memory all the pre- and post-activations in addition to the weights. 

We remark that the intermediate noise generative process admits the classical generative process \eqref{eq:classical_gen} and the SFNN generative model as special cases. Setting all $\Delta$s (and hence all $\epsilon$) to zero in \eqref{eq:gen_process_intermediate_noise} except for $\Delta_Z^{(L+1)}$ indeed gives back the classical generative process \eqref{eq:classical_gen}, with $\ell(y,z)=(y-z)^2$ and $\Delta=\Delta_Z^{(\ell+1)}$. Instead, setting $\Delta_X^{(\ell)}=0$ for all $\ell$, but keeping the noise in the pre-activations gives the SFNN model.

\vspace{-5mm}
\section{Gibbs sampler for neural networks}
\vspace{-3mm}
\label{sec:gibbs_sampler}
Gibbs sampling \cite{geman1984stochastic} is an MCMC algorithm that updates each variable in sequence by sampling it from its conditional distribution. For a probability measure with three variables $P(\theta_1,\theta_2,\theta_3)$, one step of Gibbs sampling can be described as follows. Starting from the configuration $\theta_1(t),\theta_2(t),\theta_3(t)$, we first draw $\theta_1 (t+1)\sim P(\theta_1|\theta_2(t),\theta_3(t))$, then we draw $\theta_2(t+1)\sim P(\theta_2|\theta_1(t+1),\theta_3(t))$ and finally $\theta_3(t+1)\sim P(\theta_3|\theta_1(t+1),\theta_2(t+1))$. Repeating this procedure one can prove \cite{casella1992explaining,robert1999monte} that, in the limit of many iterations ($ t\gg 1$) and provided that the chain is ergodic, the samples $(\theta_1(t),\theta_2(t),\theta_3(t))$ will come from $P(\theta_1,\theta_2,\theta_3)$. 
We now present a Gibbs sampler for the intermediate noise posterior \eqref{eq:intermediate_noise_posterior}, with Gaussian prior. More specifically the prior on $W^{(\ell)}$ is i.i.d. $\mathcal{N}(0,1/\lambda_W^{(\ell)})$ over the weights' coordinates. The full derivation of the algorithm is reported in Appendix \ref{app:Gibbs_derivation_mlp}, here we sketch the main steps. To define the sampler we need to compute the distributions of each of $X^{(\ell)}, Z^{(\ell)}, W^{(\ell)}$ conditioned on all other variables (here indicated by "All"). For $X^{(\ell)}$ the conditional distribution factorizes over samples $\mu\in[n]$, leading to
\begin{align}
\label{eq:cond_X}
&P(X^{(\ell)\mu}|\text{All})=P(X^{(\ell)\mu}|Z^{(\ell)\mu},W^{(\ell)},Z^{(\ell+1)\mu})=\mathcal N(X^{(\ell)\mu}|(m^{(X_\ell)})^\mu,\Sigma^{(X_\ell)}).
\end{align}
This is a multivariate Gaussian with covariance $\Sigma^{(X_\ell)}=\left(\frac{1}{\Delta^{(\ell+1)}_{Z}}W^{(\ell)T}W^{(\ell)}+\frac{1}{\Delta^{(\ell)}_{X}}\mathbb I_{d_\ell}\right)^{-1}$, and mean $(m^{(X_\ell)})^\mu=\Sigma^{(X_\ell)}\left(\frac{1}{\Delta^{(\ell)}_{X}}\sigma(Z^{(\ell)\mu})+\frac{1}{\Delta^{(\ell+1)}_{Z}}W^{(\ell)T}Z^{(\ell+1)\mu}\right).$

Considering $W^{(\ell)}$, we exploit that the conditional factorizes over the rows $\alpha\in[d_{\ell+1}]$.
\begin{align}
\label{eq:cond_W}
&P(W^{(\ell)}_\alpha|\text{All})=P(W^{(\ell)}_\alpha|X^{(\ell)},Z^{(\ell+1)}_\alpha)=\mathcal N(W_\alpha^{(\ell)}|(m_{W}^{(\ell)})_\alpha,\Sigma^{(\ell)}_W),
\end{align}
with $\Sigma^{(\ell)}_W=\left(\frac{1}{\Delta^{(\ell+1)}_{Z}} X^{(\ell)T}X^{(\ell)}+\lambda^{(\ell)}_W\mathbb I_{d_\ell}\right)^{-1}$, and $(m^{(\ell)}_{W})_\alpha=\frac{1}{\Delta^{(\ell+1)}_{Z}}\Sigma^{(\ell)}_W X^{(\ell)T}Z^{(\ell+1)}_\alpha.$

For $Z^{(\ell+1)}$ the conditional factorizes both over samples and over coordinates. We have
\begin{align}
\label{eq:cond_Z}
&P(Z^{(\ell+1)\mu}_\alpha|\text{All})=P(Z^{(\ell+1)\mu}_\alpha|X^{(\ell+1)\mu}_\alpha,W^{(\ell)}_\alpha, X^{(\ell)\mu})\propto\\&\exp\left[-\frac{1}{2\Delta^{(\ell+1)}_{Z}}\left(Z^{(\ell+1)\mu}_\alpha-W^{(\ell)T}_\alpha X^{(\ell)\mu}\right)^2-\frac{1}{2\Delta^{(\ell+1)}_{X}}\left(\sigma(Z^{(\ell+1)\mu}_\alpha)-X^{(\ell+1)\mu}_\alpha\right)^2\right].\nonumber
\label{eq:Z_conditional}
\end{align}

Notice that the conditional distributions of $W^{(\ell)}_\alpha$ and $X^{(\ell)\mu}$ are multivariate Gaussians and can be easily sampled. Instead $Z^{(\ell)\mu}_\alpha$ is a one-dimensional random variable with non Gaussian distribution. Appendix \ref{app:sample_Z} provides recipes for sampling it for sign, ReLU and absolute value activations. 

\begin{algorithm}[H]
\caption{Gibbs sampler for Multilayer perceptron} 
\label{alg:Gibbs sampler}
\begin{algorithmic}
\State\textbf{Input: } training inputs $X$, training labels $y$, noise variances $\{\Delta_Z^{(\ell)}\}_{\ell=2}^{L+1}$, $\{\Delta_X^{(\ell)}\}_{\ell=2}^L$, prior inverse variances $\{\lambda_W^{(\ell)}\}_{\ell=1}^L$, initial condition $\{X^{(\ell)}\}_{\ell=2}^L$,$\{W^{(\ell)}\}_{\ell=1}^L$, $\{Z^{(\ell)}\}_{\ell=2}^{L}$, length of the simulation $t_\text{max}$
\State\textbf{Output:} a sequence $S$ of samples
\State $X^{(1)}\gets X$
\State $Z^{(L+1)}\gets y$
\State $S\gets \left[(\{W^{(\ell)}\}_{\ell=1}^L,\, \{X^{(\ell)}\}_{\ell=2}^L,\,\{Z^{(\ell)}\}_{\ell=2}^{L})\right]$

\For{$t=1,\dots,t_\text{max}$}
    \State $W^{(1)} \sim P(W^{(1)}|X,Z^{(2)})$ \Comment{See eq.~\eqref{eq:cond_W}}
    \For{$\ell=2,\dots,L$}
    \State $X^{(\ell)} \sim P(X^{(\ell)}|Z^{(\ell)},W^{(\ell)},Z^{(\ell+1)})$\Comment{See eq.~\eqref{eq:cond_X}}
    \State $W^{(\ell)} \sim P(W^{(\ell)}|X^{(\ell)},Z^{(\ell+1)})$ \Comment{See eq.~\eqref{eq:cond_W}}
    \State $Z^{(\ell)} \sim P(Z^{(\ell)}|X^{(\ell-1)},W^{(\ell-1)},X^{(\ell)})$ \Comment{See eq.~\eqref{eq:cond_Z}}
    \EndFor
    \State $S.\text{append}\left( \left(\{W^{(\ell)}\}_{\ell=1}^L,\, \{X^{(\ell)}\}_{\ell=2}^L,\,\{Z^{(\ell)}\}_{\ell=2}^{L}\right)\right)$
\EndFor
\end{algorithmic}
\end{algorithm}

Putting all ingredients together, we obtain the Gibbs sampling algorithm, whose pseudocode is reported in Algorithm \ref{alg:Gibbs sampler}. The main advantages of Gibbs sampling lie in the fact that it has no hyperparameters to tune and, moreover, it is a rejection-free sampling method. In the case of MCMCs, hyperparameters are defined to be all parameters that can be changed without affecting the probability measure that the MCMC asymptotically samples. The Gibbs sampler can also be parallelized across layers: a parallelized version of Algorithm \ref{alg:Gibbs sampler} is presented in Appendix \ref{app:Gibbs_parallel}. Finally, one can also extend this algorithm to more complex architectures: Appendices \ref{app:biases} and \ref{app:cnn} contain respectively the update equations for biases and convolutional networks. We release an implementation of the Gibbs sampler at \url{https://github.com/SPOC-group/gibbs-sampler-neural-networks}

\section{Numerical results}
In this section we present numerical experiments to support our claims. We publish the code to reproduce these experiments at \url{https://github.com/SPOC-group/numerics-gibbs-sampling-neural-nets}
\label{sec:numerical results}
\subsection{Teacher student convergence method}
\label{sec:ts_conv_numerical}

In section \ref{sec:TS_sec} we proposed a thermalization criterion based on having access to an already thermalized initialization. Here we show that it is more discriminative than other commonly used heuristics. We first briefly describe these heuristics.
\begin{itemize}
    \item \textbf{Stationarity}. Thermalization implies stationarity since once the MCMC has thermalized, it samples from a fixed probability measure. Therefore any observable, plotted as a function of time should oscillate around a constant value. The converse (stationarity implies thermalization) is not true. Nevertheless observing when a function becomes stationary gives a lower bound on $T_\text{therm}$.
    \item \textbf{Score method} \cite{fan2006output}. Given a probability measure $P(W)$, we exploit the fact that $\E_{W\sim P}\left[\frac{\partial \log P(W)}{\partial W}\right]=\int \frac{\partial P(W)}{\partial W} dW=0$. We then monitor the function $\frac{\partial \log P(W)}{\partial W}$ along the dynamics. The time at which it starts fluctuating around zero is another lower bound to $T_\text{therm}$. 
    \item \textbf{ $\hat R$ statistic} \cite{gelman1992inference}. Two (or more) MCMCs are run in parallel starting from different initializations. The within-chain variance is compared to the total variance, obtained by merging samples from both chains. 
    Call the ratio of these variances $\hat R$ (a precise definition of which is given in Appendix \ref{app:Rhat}).
    If the MCMC has thermalized, the samples from the two chains should be indistinguishable, thus $\hat R$ will be close to 1. 
    The time at which $\hat{R}$ gets close to 1 provides yet another lower bound to the thermalization time.
\end{itemize}  
We compare these methods in the case of a one hidden layer neural network, identical for the teacher and the student, with input dimension $d_1=50$, $d_2=10$ hidden units and a scalar output. This corresponds to the function
\begin{equation}
    \label{eq:basic_2_layer_nn}
    f(x,W)=b^{(2)}+W^{(2)}\sigma\left(W^{(1)}x + b^{(1)}\right),
\end{equation}
 where $\sigma(x)=\max(0,x)$ and $W$ indicates the collection of all parameters:  $W^{(1)}\in\mathbb R^{d_2\times d_1}$ and $W^{(2)}\in\mathbb R^{1\times d_2}$. We specify the prior by setting $\lambda_W^{(1)}=\lambda_b^{(1)}=d_1, \lambda_W^{(2)}=\lambda_b^{(2)}=d_2$, the prior on the bias $b^{(\ell)}$ is $\mathcal{N}(0,1/\lambda^{(\ell)}_b)$ i.i.d. over the coordinates of the bias vector.
Let $n=2084$ be the size of the training set. We pick $n$ to be four times the number of parameters in the network anticipating that the training set contains enough information to learn the teacher. We start by generating the matrix of training inputs $X\in\mathbb R^{n\times d_1}$ with i.i.d. standard Gaussian entries, then we sample the teacher's weights $W_\star$ from the Gaussian prior. For concreteness we set $\Delta_Z^{(2)},\Delta_X^{(2)},\Delta_Z^{(3)}$ to the same value $\Delta$ and set $\Delta=10^{-4}$.
To generate the training labels $y$, we feed $X$, the teacher's weights $W_\star$ and $\Delta$ into the generative process \eqref{eq:gen_process_intermediate_noise}, adapted to also add the biases. 
For the test set, we first sample $X_\text{test}$, with i.i.d. standard Gaussian entries. Both the test labels and the test predictions are generated in a noiseless way (i.e., just passing the inputs through the network). In this way, the test mean square error (MSE) takes the following form: $\text{test MSE} =\frac{1}{n_\text{test}}\sum_{\mu=1}^{n_\text{test}} \left(f(X_\text{test}^\mu,W_\star)-f(X_\text{test}^\mu,W)\right)^2.$ The full details about this experiment set are in Appendix \ref{app:ts_crit_exp}.
We run the Gibbs sampler on the intermediate noise posterior starting from three different initializations: informed, zero and random. Respectively the student's variables are initialized to the teacher's counterparts, to zero, or are sampled from the prior. In this particular setting, the Gibbs sampler initialized at zero manages to thermalize, while the random initializations fail to do so. Two independent random initializations are shown, in order to be able to use the multiple chains method. 

\begin{figure}
    \centering
    \includegraphics[width=1\linewidth]{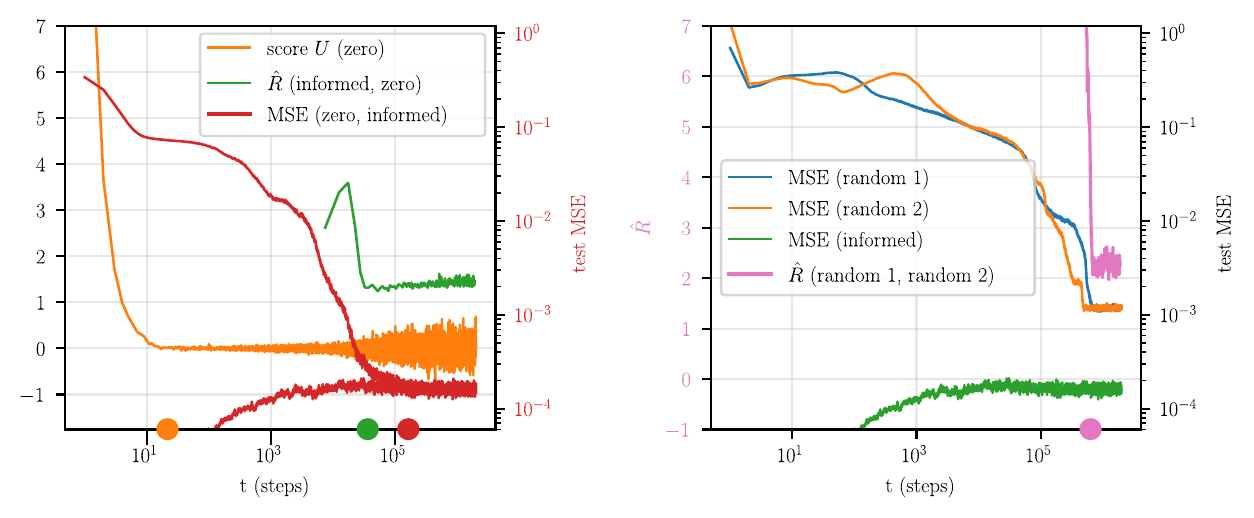}
    \caption{Comparison of different thermalization measures. In the legend, next to each method we write between parentheses the initialization (or pair of initializations) the method is applied to. The circles on the $x$ axis represent the thermalization times estimated by each method.
    \textbf{Left:} We compare the predictions for the thermalization time of the zero-initialized MCMC. The red $y$ scale on the right refers uniquely to the lines in red. All the other quantities should be read on the black $y$ scale. \textbf{Right:} We compare the predictions for the thermalization time of two chains initialized independently at random. The pink $y$ scale refers uniquely to the pink line. All other quantities should be read on the black logarithmic scale. The randomly initialized runs fail to thermalize and their test MSEs get stuck on a plateau. However, $\hat R$, whose time series on the plateau is stationary and close to 1, fails to detect this lack of thermalization.} 
    \label{fig:convergence_measures}
\end{figure}

Figure~\ref{fig:convergence_measures} illustrates a representative result of these experiments. In the left panel, we aim to find the highest lower bound to the thermalization time of the zero-initialized chain. Looking at the score method we plot $U=\Delta\frac{1}{d_1 d_2}\sum_{i=1}^{d_1}\sum_{\alpha=1}^{d_2} \frac{\partial \log P}{\partial W^{(1)}_{\alpha i}}$, where $P$ indicates the posterior distribution; this is the score rescaled by $\Delta$ and averaged over the first layer weights. In the zero-initialized chain, $U$ starts oscillating around zero already at $t=20$. 
Then we consider the $\hat R$ statistics computed on the outputs of the two chains with zero and informed initializations. The criterion estimates that the zero-initialized chain has thermalized after $t=6\times10^4$, when $\hat R$ approaches 1 and becomes stationary. Next, we consider the teacher-student method, with the test MSE as the test function ($g$ in our previous discussion). According to this method, the MCMC thermalizes after the test MSE time series of the informed and zero-initialized chains merge, which happens around $t=10^5$. Finally, the stationarity criterion, when applied to the test MSE or to $\hat R$ gives a similar estimate for the thermalization time.
The $x$-axis of the left plot provides a summary of this phenomenology, by placing a circle at the thermalization time estimated by each method. In summary, the teacher-student method is the most conservative, but the $\hat R$ statistics-based method is also reasonable here.

The right panel of figure ~\ref{fig:convergence_measures} then shows a representative situation where thermalization is not reached yet the $\hat R$ statistics-based method would indicate it is. 
In the right panel, two randomly initialized chains, denoted by \textit{random 1} and \textit{random 2} are considered. Neither of these chains actually thermalizes, in fact looking at the test MSE time series we see that both chains get stuck on the same plateau around MSE=$10^{-3}$ and are unable to reach the MSE of the informed initialization. However, as soon as both chains reach the plateau, $\hat{R}$ quickly drops to a value close to the order of 1 and thereafter becomes stationary, mistakenly signalling thermalization. This example exposes the problem at the heart of the multiple-chain method: the method can be fooled if the chains find themselves close to each other but far from equilibrium. Similarly, since the chains become stationary after they hit the plateau, the stationarity criterion would incorrectly predict that they have thermalized. To conclude, we have shown an example where common thermalization heuristics fail to recognize that the MCMC has not thermalized; instead, the teacher-student method detects the lack of thermalization.

\subsection{Gibbs sampler}
\label{sec:numerics_gibbs}
In this section, we show that the combination of intermediate noise posterior and Gibbs sampler is effective in sampling from the posterior by comparing it to HMC, run both on the classical and intermediate noise posteriors, and to MALA, run on the classical posterior. We provide the pseudocode for these algorithms in Appendix \ref{app:other_mcmc}. Notice we don't compare the Gibbs sampler to variational inference methods or altered MCMCs, since these algorithms only sample from approximated versions of the posterior. For the first set of experiments, we use the same network architecture as in the previous section. The teacher weights $W_\star$, as well as $X,X_\text{test}$ are also sampled in the same way. The intermediate noise and the classical generative process prescribe different ways of generating the labels. However, to perform a fair comparison, we use the same dataset for all MCMCs and posteriors; thus we generate the training set in a noiseless way, i.e., setting $y^\mu=f(X^\mu,W_\star)$. We generate 72 datasets according to this procedure, each time using independently sampled inputs and teacher's weigths. The consequence of generating datasets in a noiseless way is that the noise level used to generate the data is different from the one in the MCMC, implying that the informed initialization will not exactly be a sample from the posterior. However, the noise is small enough that we did not observe any noticeable difference in the functioning of the teacher-student criterion. 

First, we aim to characterize how often each algorithm thermalizes, when started from an uninformed initialization. Uninformed means that the network's initialization is agnostic to the teacher's weights. 
For several values of $\Delta$, and for all the 72 datasets, we run the four algorithms (Gibbs, classical HMC, intermediate HMC, classical MALA) starting from informed and uninformed initializations. More information about the initializations and hyperparameters of these experiments is contained in Appendix \ref{app:synth_data_exp}.

\begin{figure}
    \centering
    \includegraphics[width=0.99\linewidth]{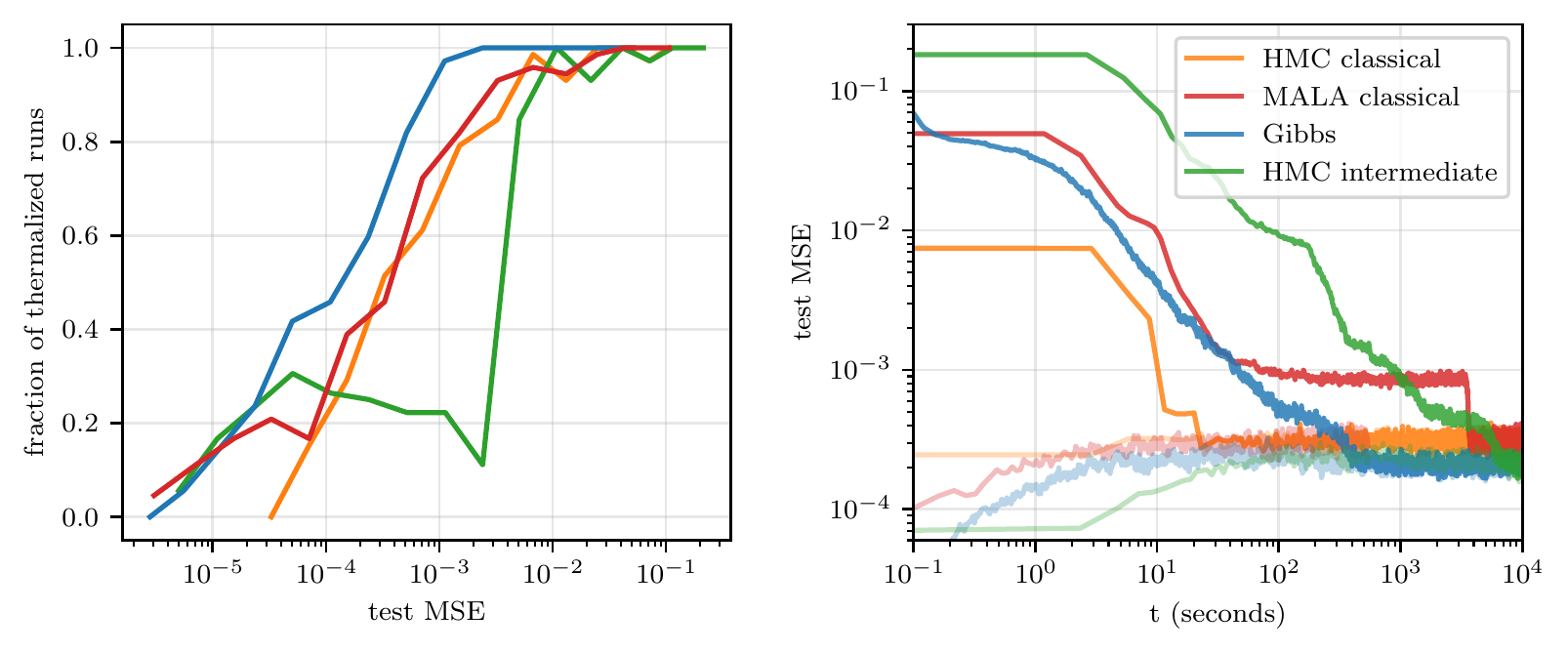}
    \caption{Thermalization experiments on synthetic data. \textbf{Left:} Proportion of the 72 runs that thermalize plotted against the equilibrium test MSE. \textbf{Right:} Example of the dynamics of the test MSE in a particular run where all four algorithms thermalize. In order to get a similar equilibrium test MSE in the classical and intermediate noise posteriors, we pick respectively $\Delta=10^{-3}$ and $\Delta=4.64\times10^{-4}$. The transparent lines represent the informed initializations.}
    \label{fig:therm_synth}
\end{figure}

The left panel of figure \ref{fig:therm_synth} depicts the proportion of the 72 datasets in which the uninformed initialization thermalizes within 5:30h of simulation. The $x-$axis is the equilibrium test MSE, i.e., the average test MSE reached by the informed initialization once it becomes stationary. When $\Delta$, and thus the test MSE, decreases, the proportion of thermalized runs drops for all algorithms, with the Gibbs sampler attaining the highest proportion, in most of the range. In the right panel, we plot the dynamics of the test error under each algorithm for a run where they all thermalize. For the same $\Delta$s of this plot (respectively $\Delta=10^{-3},4.64\times 10^{-4}$ for the classical and intermediate noise posterior), we compute the average thermalization time among the runs that thermalize. Classical HMC, MALA, Gibbs, and intermediate HMC take, respectively on average around $130, 2700,3200,12500$ seconds to thermalize. This shows that the classical HMC, when it thermalizes, is the fastest method, while MALA and Gibbs occupy the second and third position, with similar times. However classical HMC thermalizes about 20\% less often than the Gibbs sampler. Therefore in cases where it is essential to reach equilibrium, the Gibbs sampler represents the best choice.

\begin{figure}
    \centering
    \includegraphics[width=1.\linewidth]{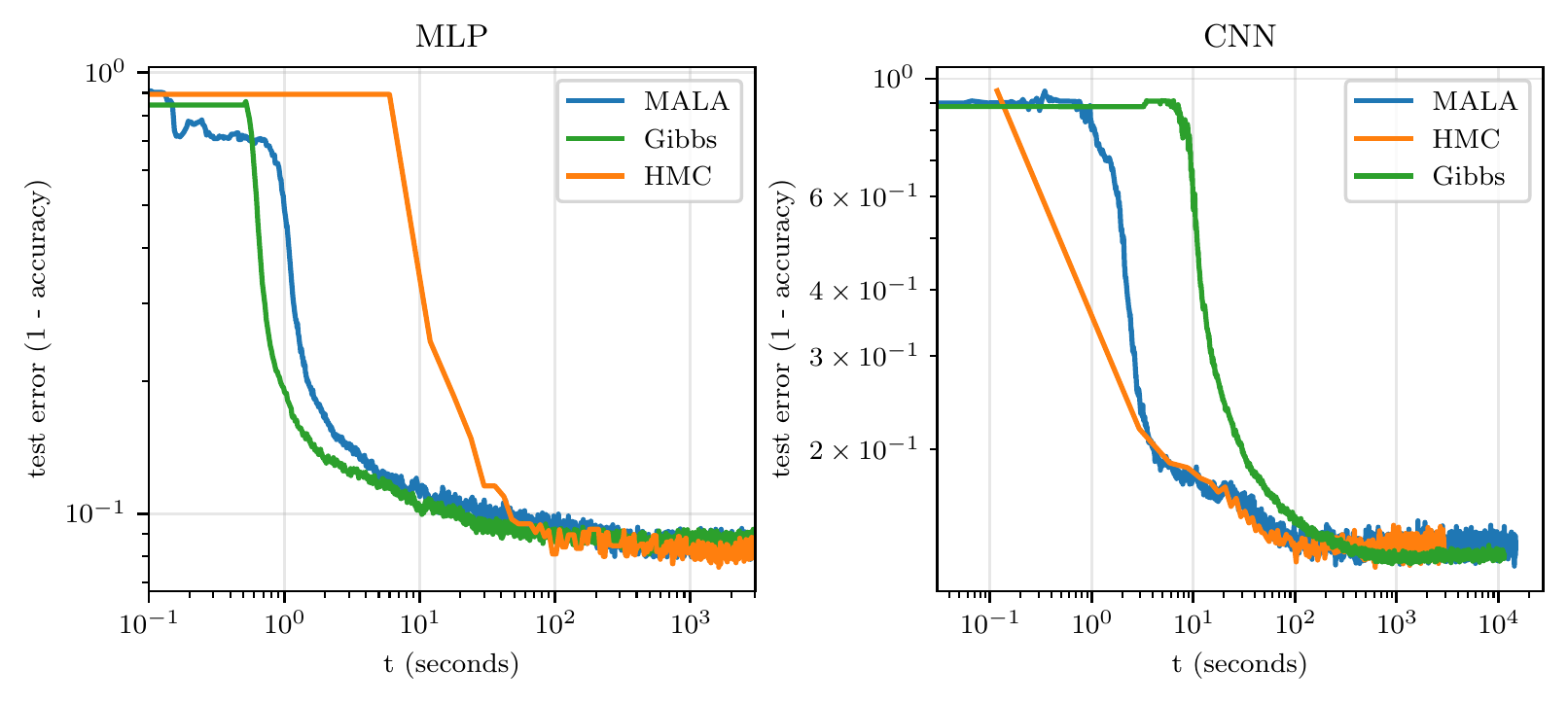}
    \caption{Gibbs on the intermediate noise posterior and HMC, MALA both on the classical posterior, compared on MNIST. \textbf{Left: } MLP with one hidden layer with $12$ hidden units. \textbf{Right:} CNN network.}
    \label{fig:mnist_figure}
\end{figure}

\label{sec:real_data_experiments}
We now move from the abstract setting of Gaussian data to more realistic inputs and architectures. As an architecture we use a one-hidden layer multilayer perceptron (MLP) with 12 hidden units and ReLU activations, and a simple convolutional neural network (CNN) with a convolutional layer, followed by average pooling, ReLU activations, and a fully connected layer. See Appendix \ref{app:real_data_experiments} for a description of both models and of the experimental details. In this setting, we resort to the stationarity criterion to check for thermalization, since the teacher-student method is inapplicable. We compare the Gibbs sampler with HMC and MALA both run on the classical posterior, picking MNIST as dataset. Figure~\ref{fig:mnist_figure} shows the test error as a function of time for the two architectures. We choose the algorithms $\Delta$s such that they all reach a comparable test error at stationarity. We then compare the time it takes each algorithm to reach this error. The results of the experiments are depicted in figure \ref{fig:mnist_figure}. For the MLP all algorithms take approximately the same time to become stationary, around $500s$. In the CNN case, after HMC and MALA reach stationarity in $100s$, compared to $800s$ for Gibbs. We however note that for HMC and MALA to achieve these performances we had to carry out an extensive optimization over hyperparameters, thus the speed is overall comparable.

\section{Conclusion}
\label{sec:discussion}
In this work, we introduced the intermediate noise posterior, a probabilistic model for Bayesian learning of neural networks, along with a novel Gibbs sampler to sample from this posterior. We compared the Gibbs sampler to MALA and HMC, varying also the form of the posterior. We found that HMC and MALA both on the classical posterior and Gibbs, on the intermediate noise posterior, each have their own merits and can be considered effective in sampling the high dimensional posteriors arising from Bayesian learning of neural networks. For the small architectures considered, Gibbs compares favourably to the other algorithms in terms of the ability to thermalize, moreover, no hyperparameter tuning is required, it can be applied to non-differentiable posteriors, and can be parallelized across layers. The main drawback of the Gibbs sampler lies in the need to store and update all the pre- and post- activations. This slows down the algorithm, compared to HMC, in the case of larger architectures.

We further proposed the teacher-student thermalization criterion: a method to obtain stringent lower bounds on the thermalization time of an MCMC, within a synthetic data setting. We first provided a simple theoretical argument to justify the method and subsequently compared it to other thermalization heuristics, finding that the teacher-student criterion consistently gives the highest lower bound to $T_\text{therm}$. 

\section{ Acknowledgments}
We thank Lucas Clarté for introducing us to the blocked Gibbs sampler, and Christian Keup for the useful discussions on predictive coding and stochastic neural networks.
This research was supported by the NCCR MARVEL, a National Centre of Competence in Research, funded by the Swiss National Science Foundation (grant number 205602).
\newpage
\bibliography{biblio.bib}
\newpage

\appendix
\section{Multiple chain convergence method (or \texorpdfstring{$\hat R$}{sigma2} statistic)}
\label{app:Rhat}
In this appendix, we recall the details of the $\hat R$ statistic, first introduced in \cite{gelman1992inference} and \cite{brooks1998general}. The method is based on running $M$ parallel MCMCs. Let $\{\theta_{mn}: m\in [M],n\in [N]\}$ be the Markov chains states. $\theta_{mn}$ here indicates the state of the $m$-th chain after $n$ steps.
The $\hat R$ statistics is formulated in terms of an arbitrary function $\psi(\theta)$. We define $\hat R$ in the following way
\begin{align}
&\bar \psi_{m \cdot} \coloneqq
\frac{1}{N}\sum_n \psi(\theta_{mn});
\quad \bar \psi_{\cdot \cdot} \coloneqq
\frac{1}{MN}\sum_{m,n} \psi(\theta_{mn});
\\
&\frac{B}{N}\coloneqq \frac{1}{M-1}\sum_m
(\bar\psi_{m\cdot} - \bar\psi_{\cdot\cdot})^2;
\\
&W\coloneqq\frac{1}{M(N-1)}\sum_{m,n}
(\psi(\theta_{mn})-\bar\psi_{m\cdot})^2;
\\
&\hat\sigma^2_+ \coloneqq \frac{N-1}{N}W + \frac{B}{N};
\\
&\hat R\coloneqq \frac{M+1}{M}\frac{\hat\sigma^2_+}{W}
- \frac{N-1}{MN}.
\end{align}
$W$ estimates the within-chain variance (averaged over all chains).
$\hat\sigma^2_+$ instead is the unbiased estimator for the variance, obtained by pooling all samples from different chains. $\hat R$ is essentially $\frac{\hat\sigma^2_+}{W}$, apart from factors that vanish when $N,M\to \infty$.
If the chains are far apart from each other then $\hat\sigma^2_+\gg W$, and hence $\hat R \gg 1$. Instead, if all chains are close to each other then $\hat R$ will be close to 1. $\hat R$ can be used to derive a lower bound to the thermalization time: when all $M$ chains have thermalized $\hat R$ should be close to 1, as samples from different chains should be indistinguishable, and thus have the same variance as samples from a single chain.

\section{Teacher-student criterion }
\label{app:ts_crit_exp}

\subsection{Properties of the informed initialization}
In this paragraph we look more in detail at the properties of the informed initialization. We saw in section \ref{sec:TS_sec} that $W_\star$ is a sample from the posterior $P(W|D)$. What does this imply for the chain initialized at $W_\star$? First any observable $\varphi(\cdot)$ that concentrates under the posterior and that does not depend explicitly on $W_\star$ (e.g. $\varphi(W)=||W||$), will be thermalized already at $t=0$.This implies that all these observables will be stationary from the very beginning of the MCMC simulation and they will oscillate around their mean value under the posterior. 
The case where $\varphi(\cdot)$ depends explicitly on $W_\star$ is more delicate. We comment on it because the observable we use to determine thermalization throughout the whole paper is the test MSE ($\varphi_\text{MSE}(W)=\frac{1}{n_\text{test}}\sum_{\mu=1}^{n_\text{test}} \left(f(X_\text{test}^\mu,W_\star)-f(X_\text{test}^\mu,W)\right)^2$), which explicitly depends on $W_\star$. In the following we will explore the behavior of the test MSE, however keep in mind that most other observables dependent on $W_\star$ will exhibit similar behavior. 
The first peculiarity of the test MSE, is that under the informed initialization it is not stationary, and it is not close to its expected value under the posterior. In fact at initialization we always have test MSE=0. As more samples are drawn the MSE then relaxes to equilibrium and starts oscillating around its expected value under the posterior. If the test MSE is not thermalized, what is then the advantage of using the informed initialization compared to an uninformed one?
While the test MSE is not thermalized, most other observables are under the informed initialization. This means that the chain is started in a favorable region of the weight space. In practice, looking at the right panel of figure \ref{fig:therm_synth} one can compare the smooth convergence to stationarity of the informed initializations (transparent lines), to the irregular paths followed by the uninformed initializations (solid lines). 

\subsection{Numerical experiments}
In the rest of this appendix, we report the details of the experiments presented in \ref{sec:ts_conv_numerical}. Recall that a synthetic dataset was generated using a teacher network with Gaussian weights, and according to the intermediate noise generative process \eqref{eq:gen_process_intermediate_noise}. Then the Gibbs sampler was used to sample from the resulting posterior. We precise that all parameters of the Gibbs sampler (i.e. all the $\Delta$s and $\lambda$s) match those of the generative process. The Gibbs sampler was run on four chains: one with informed initialization, one initialized at zero, and two chains with independent random initializations. For the random initializations the weights of the student are sampled from the prior (with the same $\lambda$s as the teacher), then the pre- and post-activations are computed using the intermediate noise generative process \eqref{eq:gen_process_intermediate_noise}, with noises $\epsilon$, independent from those of the teacher. The zero-initialized chain plausibly thermalizes, while the randomly initialized ones do not. We briefly comment on how the $\hat R$ statistic and the score statistics were computed.

\subsection[Rhat statistic]{\texorpdfstring{$\hat R$ }{} statistic}
\begin{figure}
    \centering
    \includegraphics[width=1\linewidth]{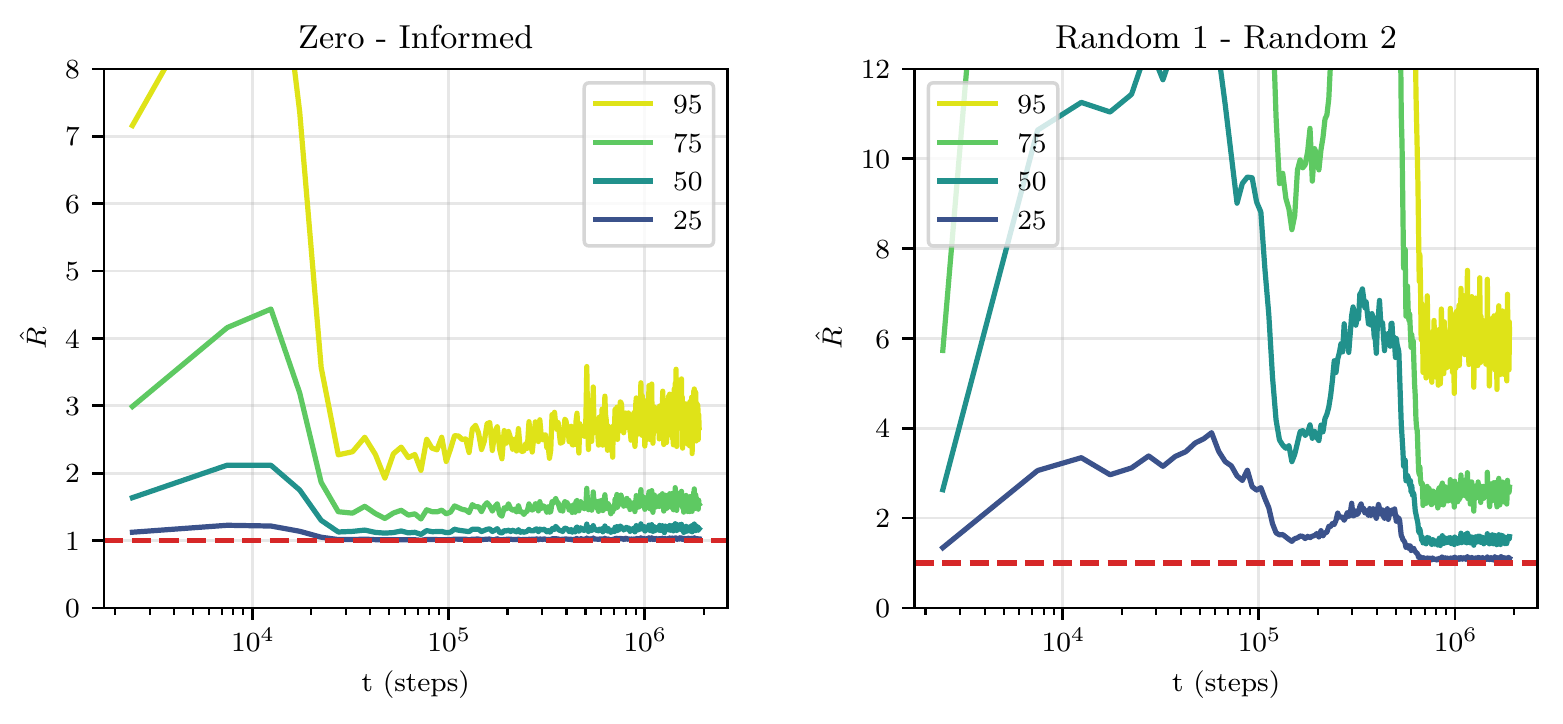}
    \caption{Percentiles of $\hat R$ as a function of time. A line marked with the number $k$ in the legend represents how the $k$-th percentile of $\hat R$ changes throughout the simulation. 
    The data comes from the same simulation that was used for computing the average $\hat R$ in figure \ref{fig:convergence_measures}. 
    The red dashed horizontal line is placed at a height of 1, the value that $\hat R$ should approach when the chains are close to each other.
    \textbf{Left:} percentiles of $\hat R$, computed on two chains with respectively zero and informed initialization. \textbf{Right:} percentiles when computing $\hat R$ from two chains independently initialized at random.}
    \label{fig:quantiles_Rhat}
\end{figure}

In the notation of \ref{app:Rhat},  $\theta$ is given by $\{W^{(1)},b^{(1)},Z^{(2)},X^{(2)},W^{(2)},b^{(2)}\}$.
Next we have to pick the  (possibly vector-valued) function $\psi(\theta)$. One possible choice is to use the weights, e.g., $\psi(\theta)=W^{(1)}$. However, due to the permutational symmetry between the neurons in the hidden layer, this gives $\hat R\gg 1$ even when the MCMC has thermalized. Hence one must focus on quantities that are invariant to this symmetry. A natural choice is the student output on the test set. We pick $\psi(\theta)\in \mathbb R^{n_\text{test}}$, with  $\psi(\theta)=f(X_\text{test},W)$ and $f$ as in \eqref{eq:basic_2_layer_nn}. We record these vectors along the simulation at times evenly spaced by 100 MCMC steps. We split the samples into blocks of 50 consecutive measurements. We then compute the $\hat R$ statistic on each block (hence $N=50$). Since the function $\psi$ we are using returns an $n_\text{test}$ dimensional output, $\hat R$ will also be $n_\text{test}$ dimensional. In figure \ref{fig:convergence_measures} we then decided to plot the average (over the test set) value of $\hat R$. In other words, calling $\hat R^\nu_\tau$ the value of $\hat R$ computed on the $\tau-$th block and the $\nu$-th test sample, what we plot are the pairs $\left(t_\tau,\frac{1}{n_\text{test}}\sum_{\nu=1}^{n_\text{test}}\hat R^\nu_\tau\right)$, with $t_\tau$ being the average time within block $\tau$.
In principle the whole distribution of $\hat R$ is interesting. Figure \ref{fig:quantiles_Rhat} shows the evolution of the 25th, 50th, 75th and 95th percentiles of $\hat R$.
Recall that the two chains in the left panel thermalize, while those in the right panel are actually very far from equilibrium. Even if the distribution in the right panel is more shifted away from one than the distribution in the left panel, we still think that the sudden drop from a higher value and the subsequent stationarity could be interpreted as the chains having thermalized. 
\subsection{Score method}
If a MCMC has thermalized then the gradient of the log posterior must have mean zero, hence the time series of each of its coordinates will have to oscillate around zero. In figure \ref{fig:convergence_measures} we plot $U=\Delta\frac{1}{d_1 d_2}\sum_{i=1}^{d_1}\sum_{\alpha=1}^{d_2} \frac{\partial \log P}{\partial W^{(1)}_{\alpha i}}$, where $P$ indicates the posterior distribution; this is the score rescaled by $\Delta$ and averaged over the first layer weights. We also tried taking the gradient with respect to the second layer weights $W^{(2)}$, or with respect to $Z^{(2)}$. The results do not change significantly and the score methods keep severely underestimating the thermalization time.
\section{Gibbs sampler derivation}
\label{app:Gibbs_derivation_mlp}
In this appendix, we provide the details of the derivation of the Gibbs sampler algorithm in the case of an MLP without biases. In order, we will derive equations \eqref{eq:cond_X}, \eqref{eq:cond_W} and \eqref{eq:cond_Z}.

For $X^{(\ell)}$ we see that the conditional distribution factorizes over samples $\mu\in[n]$.
\begin{align}\nonumber
&P(X^{(\ell)\mu}|\text{All})=P(X^{(\ell)\mu}|Z^{(\ell)\mu},W^{(\ell)},Z^{(\ell+1)\mu})=\frac{P(X^{(\ell)\mu}, Z^{(\ell+1)\mu}|W^{(\ell)},Z^{(\ell)\mu})}{P(Z^{(\ell)\mu}|W^{(\ell)},Z^{(\ell)\mu})}=\\&\nonumber=\frac{P(Z^{(\ell+1)\mu}|W^{(\ell)},X^{(\ell)\mu})P(X^{(\ell)\mu}|W^{(\ell)},Z^{(\ell)\mu})}{P(Z^{(\ell)\mu}|W^{(\ell)},Z^{(\ell-1)\mu})}\propto P(Z^{(\ell+1)\mu}|W^{(\ell)},X^{(\ell)\mu})P(X^{(\ell)\mu}|Z^{(\ell)\mu})\\&=\mathcal N(Z^{(\ell+1)\mu}|W^{(\ell)}X^{(\ell)\mu},\Delta^{(\ell+1)}_{Z}\mathbb{I}_{d_{\ell+1}})\,\mathcal N(X^{(\ell)\mu}|\sigma(Z^{(\ell)\mu}),\Delta^{(\ell)}_{X}\mathbb I_{d_{\ell}})=\mathcal N(X^{(\ell)\mu}|m^{(\ell)\mu}_X,\Sigma^{(\ell)}_X),
\end{align}
with $\Sigma^{(X_\ell)}=\left(\frac{1}{\Delta^{(\ell+1)}_{Z}}W^{(\ell)T}W^{(\ell)}+\frac{1}{\Delta^{(\ell)}_{X}}\mathbb I_{d_\ell}\right)^{-1}$, and $(m^{(X_\ell)})^\mu=\Sigma^{(X_\ell)}\left(\frac{1}{\Delta^{(\ell)}_{X}}\sigma(Z^{(\ell)\mu})+\frac{1}{\Delta^{(\ell+1)}_{Z}}W^{(\ell)T}Z^{(\ell+1)\mu}\right).$ In the derivation we used that $P(X^{(\ell)}|W^{(\ell)},Z^{(\ell)})=P(X^{(\ell)}|Z^{(\ell)})$.
We see that the conditional distribution of $X^{(\ell)}$ is a multivariate Gaussian, hence it is possible to sample from it efficiently on a computer.

\begin{algorithm}[H]
\caption{Parallel Gibbs sampler for MLP. The expression \textbf{for parallel} indicates that all iterations of the loop can be executed in parallel.} 
\label{alg: parallel Gibbs sampler}
\begin{algorithmic}
\State\textbf{Input: } training inputs $X$, training labels $y$, noise variances $\{\Delta_Z^{(\ell)}\}_{\ell=2}^{L+1}$, $\{\Delta_X^{(\ell)}\}_{\ell=2}^L$, prior inverse variances $\{\lambda_W^{(\ell)}\}_{\ell=1}^L$, initial condition $\{X^{(\ell)}\}_{\ell=2}^L$,$\{W^{(\ell)}\}_{\ell=1}^L$, $\{Z^{(\ell)}\}_{\ell=2}^{L}$, length of the simulation $t_\text{max}$
\State\textbf{Output: } a sequence $S$ of samples
\State $X^{(1)}\gets X$
\State $Z^{(L+1)}\gets y$
\State $S\gets [\{W^{(\ell)}\}_{\ell=1}^L,\, \{X^{(\ell)}\}_{\ell=2}^L,\,\{Z^{(\ell)}\}_{\ell=2}^{L}]$
\For{$t=1,\dots,t_\text{max}$}
    \For{ \textbf{parallel} $\ell=2,\dots,L$}
    \State $X^{(\ell)} \sim P(X^{(\ell)}|Z^{(\ell)},W^{(\ell)},Z^{(\ell+1)})$ \Comment{\eqref{eq:cond_X}}
    \EndFor
    \For{ \textbf{parallel} $\ell=1,\dots,L$}
    \State $W^{(\ell)} \sim P(W^{(\ell)}|X^{(\ell)},Z^{(\ell+1)})$ \Comment{\eqref{eq:cond_W}}
    \EndFor
    \For{ \textbf{parallel} $\ell=2,\dots,L$}
    \State $Z^{(\ell)} \sim P(Z^{(\ell)}|X^{(\ell-1)},W^{(\ell-1)},X^{(\ell)})$ \Comment{\eqref{eq:cond_Z}}
    \EndFor
    \State $S.\text{append}\left( \{W^{(\ell)}\}_{\ell=1}^L,\, \{X^{(\ell)}\}_{\ell=2}^L,\,\{Z^{(\ell)}\}_{\ell=2}^{L}\right)$ 
\EndFor
\end{algorithmic}
\end{algorithm}

For $W^{(\ell)}$ we exploit that the conditional factorizes over the rows $\alpha\in[d_{\ell+1}]$.
\begin{align}
&P(W^{(\ell)}_\alpha|\text{All})=P(W^{(\ell)}_\alpha|X^{(\ell)},Z^{(\ell+1)}_\alpha)=\frac{P(Z^{(\ell+1)}_\alpha|W_\alpha^{(\ell)},X^{(\ell)})P(W_\alpha^{(\ell)}|X^{(\ell)})}{P(Z^{(\ell+1)}_\alpha|X^{(\ell)})}\\&\propto P(Z^{(\ell+1)}_\alpha|W_\alpha^{(\ell)},X^{(\ell)})P(W_\alpha^{(\ell)})=\mathcal N(Z^{(\ell+1)}_\alpha|X^{(\ell)}W^{(\ell)}_\alpha, \Delta^{(\ell+1)}_Z)\mathcal N(W^{\ell}_\alpha|0,\mathbb I_{d_\ell}/\lambda^{(\ell)}_W)\\&=\mathcal N(W_\alpha^{(\ell)}|m_{W\alpha}^{(\ell)},\Sigma^{(\ell)}_W),
\end{align}
with $\Sigma^{(\ell)}_W=\left(\frac{1}{\Delta^{(\ell+1)}_{Z}} X^{(\ell)T}X^{(\ell)}+\lambda^{(\ell)}_W\mathbb I_{d}\right)^{-1}$, and $(m^{(\ell)}_{W})_\alpha=\frac{1}{\Delta^{(\ell+1)}_{Z}}\Sigma^{(\ell)}_W X^{(\ell)T}Z^{(\ell+1)}_\alpha$
Once again the conditional distribution of $W^{(\ell)}$ is a multivariate Gaussian.\\
In the case of $Z^{(\ell+1)}$ the conditional factorizes both over samples and over coordinates. We have
\begin{align}
&P(Z^{(\ell+1)\mu}_\alpha|\text{All})=P(Z^{(\ell+1)\mu}_\alpha|X^{(\ell+1)\mu}_\alpha,W^{(\ell)}_\alpha, X^{(\ell)\mu})=\frac{P(Z^{(\ell+1)\mu}_\alpha,X^{(\ell+1)\mu}_\alpha|W^{(\ell)}_\alpha,X^{(\ell)\mu})}{P(X^{(\ell+1)\mu}_\alpha|W^{(\ell)}_\alpha,X^{(\ell)\mu})}\\&=\frac{P(Z^{(\ell+1)\mu}_\alpha|X^{(\ell)\mu}_\alpha,W^{(\ell)}_\alpha)P(X^{(\ell+1)\mu}_\alpha|Z^{(\ell+1)\mu}_\alpha)}{P(X^{(\ell+1)\mu}_\alpha|W^{(\ell)}_\alpha,X^{(\ell)\mu})}\\&\propto P(Z^{(\ell+1)\mu}_\alpha|X^{(\ell)\mu}_\alpha,W^{(\ell)}_\alpha)P(X^{(\ell+1)\mu}_\alpha|Z^{(\ell+1)\mu}_\alpha)\\&=\mathcal N(Z^{(\ell+1)\mu}_\alpha|W^{(\ell)T}_\alpha X^{(\ell)\mu}, \Delta^{(\ell+1)}_{Z})\mathcal N(X^{(\ell+1)\mu}_\alpha|\sigma(Z^{(\ell+1)\mu}_\alpha),\Delta^{(\ell+1)}_{X})\propto \\&\propto \exp\left[-\frac{1}{2\Delta^{(\ell+1)}_{Z}}\left(Z^{(\ell+1)\mu}_\alpha-W^{(\ell)T}_\alpha X^{(\ell)\mu}\right)^2-\frac{1}{2\Delta^{(\ell+1)}_{X}}\left(\sigma(Z^{(\ell+1)\mu}_\alpha)-X^{(\ell+1)\mu}_\alpha\right)^2\right].
\label{eq:Z_conditional_app}
\end{align}
\section{Parallelizability of the Gibbs sampler}
\label{app:Gibbs_parallel}
In this appendix, we present a version of the Gibbs sampler that is parallelized across layers. Its pseudocode is reported in algorithm \ref{alg: parallel Gibbs sampler}.

Notice that operations within each of the inner loops over $\ell$ can be executed in parallel, as there are no cross dependencies between different $\ell$s: for example, one can sample $X^{(2)},X^{(3)}\text{ and }X^{(4)}$ in parallel. This implies that, provided that we have enough computing power, the running time of the algorithm can be made independent of the depth of the network. Also updating a variable in layer $\ell$ only requires knowing the state of variables in adjacent layers (i.e. $\ell-1$, $\ell+1$), thus from a memory point of view the variables can be stored in different nodes in a cluster, with minimal communication between nodes necessary. Suppose for example that the depth of the network, $L$, is even and that we store the variables from $X$ to $Z^{(L/2)}$ in the first node, and the variables from $X^{(L/2)}$ to $y$ in the second. Then, when running the Gibbs sampler, it is only necessary to synchronize $Z^{(L/2)}$ and $X^{(L/2)}$ between the nodes. 
\section{Sampling from \texorpdfstring{$P(Z^{(\ell+1)\mu}_\alpha|X^{(\ell+1)\mu}_\alpha,W^{(\ell)}
_\alpha, X^{(\ell)\mu})$}{sigma2}}
\label{app:sample_Z}
In this section, we specify how to sample the pre-activations $Z^{(\ell)}$ in the Gibbs sampler.
Given an activation function $\sigma(\cdot)$, our goal is to sample the one dimensional random variable $Z^{(\ell+1)\mu}_\alpha$ with distribution 
\begin{align}
    P(Z^{(\ell+1)\mu}_\alpha|X^{(\ell+1)\mu}_\alpha,W^{(\ell)}_\alpha, X^{(\ell)\mu})=&\frac{1}{\mathcal Z}\exp\left[-\frac{1}{2\Delta^{(\ell+1)}_{Z}}\left(Z^{(\ell+1)\mu}_\alpha-W^{(\ell)T}_\alpha X^{(\ell)\mu}\right)^2\right.\\&\left.-\frac{1}{2\Delta^{(\ell+1)}_{X}}\left(\sigma(Z^{(\ell+1)\mu}_\alpha)-X^{(\ell+1)\mu}_\alpha\right)^2\right]
\end{align}
We shall now provide sampling algorithm for various activation functions.
\subsection[ReLU]{\texorpdfstring{$\sigma(z)=max(0,z)$}{}}
We have two cases depending on whether $Z^{(\ell+1)\mu}_\alpha$ is positive or negative.

Consider first the case $Z^{(\ell+1)\mu}_\alpha\leq 0$.

In this case $P(Z^{(\ell+1)\mu}_\alpha|X^{(\ell+1)\mu}_\alpha,W^{(\ell)}
_\alpha, X^{(\ell)\mu})\propto\mathcal N(Z^{(\ell+1)\mu}_\alpha|W^{(\ell)T}_\alpha X^{(\ell)\mu}, \Delta^{(\ell+1)}_Z)$, hence the term coming from $X^{(\ell+1)}$ does not appear. The mass of this part of the distribution is 
\begin{align}
    \mathcal Z_{-}&=\int_{-\infty}^0 \exp\left[-\frac{1}{2\Delta^{(\ell+1)}_Z} z^2+\frac{1}{\Delta^{(\ell+1)}_Z} z W^{(\ell)T}_\alpha X^{(\ell)\mu}\right] dz\\&=\sqrt{\frac{\pi \Delta^{(\ell+1)}_Z}{2}}e^{\frac{1}{2\Delta^{(\ell+1)}_Z}(W^{(\ell)T}_\alpha X^{(\ell)\mu})^2}\text{erfc}\left(\frac{W^{(\ell)T}_\alpha X^{(\ell)\mu}}{\sqrt{2\Delta^{(\ell+1)}_Z}}\right)
\end{align}
We now look at the case $Z^{(\ell+1)\mu}_\alpha>0$.
\begin{align}
&P(Z^{(\ell+1)\mu}_\alpha|X^{(\ell+1)\mu}_\alpha,W^{(\ell)}
_\alpha, X^{(\ell)\mu})\propto \mathcal N(Z^{(\ell+1)\mu}_\alpha|W^{(\ell)T}_\alpha X^{(\ell)\mu}, \Delta^{(\ell+1)}_Z)\times\\&\times\mathcal N(X^{(\ell+1)\mu}_\alpha|Z^{(\ell+1)\mu}_\alpha,\Delta^{(\ell+1)}_X)\propto\\&\propto \mathcal N\left(Z^{(\ell+1)\mu}_\alpha\bigg|\frac{\Delta^{(\ell+1)}_XW^{(\ell)T}_\alpha X^{(\ell)\mu}+\Delta^{(\ell+1)}_ZX^{(\ell+1)\mu}_\alpha}{\Delta^{(\ell+1)}_X+\Delta^{(\ell+1)}_Z}, \frac{\Delta^{(\ell+1)}_X\Delta^{(\ell+1)}_Z}{\Delta^{(\ell+1)}_X+\Delta^{(\ell+1)}_Z}\right).
\end{align}
The mass of this part is 
\begin{align}
&\mathcal Z_+=\int^{\infty}_0 \exp\left[-\frac{1}{2}\left(\frac{\Delta^{(\ell+1)}_X+\Delta^{(\ell+1)}_Z}{\Delta^{(\ell+1)}_X\Delta^{(\ell+1)}_Z}\right) z^2+z\left(\frac{X^{(\ell+1)\mu}_\alpha}{\Delta^{(\ell+1)}_X}+\frac{W^{(\ell)T}_\alpha X^{(\ell)\mu}}{\Delta^{(\ell+1)}_Z}\right)\right] dz\\&=\sqrt{\frac{\pi \Delta^{(\ell+1)}_X\Delta^{(\ell+1)}_Z}{2(\Delta^{(\ell+1)}_X+\Delta^{(\ell+1)}_Z)}}\exp\left[\frac{\left(\Delta^{(\ell+1)}_ZX^{(\ell+1)\mu}_\alpha+\Delta^{(\ell+1)}_XW^{(\ell)T}_\alpha X^{(\ell)\mu}\right)^2}{2(\Delta^{(\ell+1)}_X+\Delta^{(\ell+1)}_Z)\Delta^{(\ell+1)}_X\Delta^{(\ell+1)}_Z}\right]\times\\&\times\left[\text{erf}\left(\frac{\Delta^{(\ell+1)}_ZX^{(\ell+1)\mu}_\alpha+\Delta^{(\ell+1)}_XW^{(\ell)T}_\alpha X^{(\ell)\mu}}{\sqrt{2(\Delta^{(\ell+1)}_X+\Delta^{(\ell+1)}_Z)\Delta^{(\ell+1)}_X\Delta^{(\ell+1)}_Z}}\right)+1\right]\end{align}
Then one has $\mathcal Z=\mathcal Z_++\mathcal Z_-$.
The probability of having $Z^{(\ell+1)\mu}_\alpha<0$ is $p_-=\frac{\mathcal Z_-}{\mathcal Z}=\frac{1}{1+\frac{\mathcal{Z_+}}{\mathcal{Z_-}}}$
\subsubsection{Sampling}
First draw a bernoulli variable $r\sim\text{Bernoulli}(p_-)$.
If $r=1$ sample a negative truncated normal from the $z<0$ distribution.
If $r=0$ sample a positive truncated normal from the $z>0$ distribution.
\subsection[Sign]{\texorpdfstring{$\sigma(z)=\text{sign}(z)$}{}}
If  $Z_\alpha^{(\ell+1)\mu}>0$ (resp. $<0$) one ends up sampling from positive (resp. negative) part of the following Gaussian
\begin{align}
\label{eq:P_Z_sign}
&P(Z^{(\ell+1)\mu}_\alpha|X^{(\ell+1)\mu}_\alpha,W^{(\ell)}
_\alpha, X^{(\ell)\mu})\propto\\&\propto\frac{1}{\mathcal Z}\exp\left[\frac{1}{\Delta^{(\ell+1)}_Z}Z_\alpha^{(\ell+1)\mu} W^{(\ell)T}_\alpha X^{(\ell)\mu}-\frac{1}{2\Delta^{(\ell+1)}_Z}(Z_\alpha^{(\ell+1)\mu})^2\right]\propto\mathcal N(W^{(\ell)T}_\alpha X^{(\ell)\mu},\Delta^{(\ell+1)}_Z)
\end{align}
We now compute the normalization associated to the positive and negative parts

Positive part
\begin{align}
\mathcal Z_+&=\int_0^\infty \exp\left[\frac{1}{\Delta^{(\ell+1)}_X} X_\alpha^{(\ell+1)\mu}+\frac{1}{\Delta^{(\ell+1)}_Z}Z_\alpha^{(\ell+1)\mu} W^{(\ell)T}_\alpha X^{(\ell)\mu}-\frac{1}{2\Delta^{(\ell+1)}_Z}(Z_\alpha^{(\ell+1)\mu})^2\right]=\\&=\sqrt{\frac{\pi \Delta^{(\ell+1)}_Z}{2}}\exp\left[\frac{1}{\Delta^{(\ell+1)}_X} X_\alpha^{(\ell+1)\mu}\right]\exp\left(\frac{\left(W^{(\ell)T}_\alpha X^{(\ell)\mu}\right)^2}{2 \Delta^{(\ell+1)}_Z}\right)\left[1+\text{erf}\left(\frac{W^{(\ell)T}_\alpha X^{(\ell)\mu}}{\sqrt{2 \Delta^{(\ell+1)}_Z}}\right)\right]
\end{align}
Negative part
\begin{align}
    &\mathcal Z_-=\int_{-\infty}^0 \exp\left[-\frac{1}{\Delta^{(\ell+1)}_X} X_\alpha^{(\ell+1)\mu}+\frac{1}{\Delta^{(\ell+1)}_Z}Z_\alpha^{(\ell+1)\mu} W^{(\ell)T}_\alpha X^{(\ell)\mu}-\frac{1}{2\Delta^{(\ell+1)}_Z}(Z_\alpha^{(\ell+1)\mu})^2\right]\\&=\sqrt{\frac{\pi \Delta^{(\ell+1)}_Z}{2}}\exp\left[-\frac{1}{\Delta^{(\ell+1)}_X} X_\alpha^{(\ell+1)\mu}\right]\exp\left(\frac{\left(W^{(\ell)T}_\alpha X^{(\ell)\mu}\right)^2}{2 \Delta^{(\ell+1)}_Z}\right)\text{erfc}\left(\frac{W^{(\ell)T}_\alpha X^{(\ell)\mu}}{\sqrt{2 \Delta^{(\ell+1)}_Z}}\right)
\end{align}
Hence the probability of selecting the $Z<0$ part is
$p_-=1/\left(1+\frac{\mathcal Z_+}{\mathcal Z_-}\right)$.
$$\frac{\mathcal Z_+}{\mathcal Z_-}=\exp\left[\frac{2}{\Delta^{(\ell+1)}_X} X_\alpha^{(\ell+1)\mu}\right]\frac{\left[1+\text{erf}\left(\frac{W^{(\ell)T}_\alpha X^{(\ell)\mu}}{\sqrt{2 \Delta^{(\ell+1)}_Z}}\right)\right]}{\text{erfc}\left(\frac{W^{(\ell)T}_\alpha X^{(\ell)\mu}}{\sqrt{2 \Delta^{(\ell+1)}_Z}}\right)}$$
\subsubsection{Sampling}
First draw a Bernoulli variable $r\sim\text{Bernoulli}(p_-)$.
If $r=1$ sample a negative truncated normal from \eqref{eq:P_Z_sign}.
If $r=0$ sample a positive truncated normal from \eqref{eq:P_Z_sign}. 
\subsection[Absolute value]{\texorpdfstring{$\sigma(x)=|x|$}{}}
Both the positive and negative parts of Gaussians and we have
\begin{equation}
\label{eq:sample_Z_abs}
P(Z^{(\ell+1)\mu}_\alpha|X^{(\ell+1)\mu}_\alpha,W^{(\ell)},X^{(\ell)})\propto\begin{cases*}
        \mathcal N(m_+,1/a) & if $z\geq0$\\
        \mathcal N(m_-,1/a) & if $z<0$
    \end{cases*}
\end{equation}
with $a=\frac{\Delta^{(\ell+1)}_X+\Delta^{(\ell+1)}_Z}{\Delta^{(\ell+1)}_X\Delta^{(\ell+1)}_Z}$ and $m_{\pm}=\frac{\Delta^{(\ell+1)}_XW^{(\ell)T}_\alpha X^{(\ell)\mu}\pm \Delta^{(\ell+1)}_ZX^{(\ell+1)\mu}_\alpha}{\Delta^{(\ell+1)}_X+\Delta^{(\ell+1)}_Z}$.
We now have to compute the masses of the positive and negative parts, we have
\begin{align}
    &\mathcal Z_+=\int_0^\infty \exp\left[\frac{1}{\Delta^{(\ell+1)}_X} X_\alpha^{(\ell+1)\mu}Z_\alpha^{(\ell+1)\mu}-\frac{1}{2\Delta^{(\ell+1)}_X}(Z_\alpha^{(\ell+1)\mu})^2+\right.\\&+\left.\frac{1}{\Delta^{(\ell+1)}_Z}Z_\alpha^{(\ell+1)\mu} W^{(\ell)T}_\alpha X^{(\ell)\mu}-\frac{1}{2\Delta^{(\ell+1)}_Z}(Z_\alpha^{(\ell+1)\mu})^2\right]=\sqrt{\frac{\pi}{2a}}e^{q_+^2}\left[1+\text{erf}(q_+)\right],
\end{align}
with $a$ as before and $q_\pm=\frac{\Delta^{(\ell+1)}_XW^{(\ell)T}_\alpha X^{(\ell)\mu}\pm \Delta^{(\ell+1)}_ZX^{(\ell+1)\mu}_\alpha}{\sqrt{2(\Delta^{(\ell+1)}_X+\Delta^{(\ell+1)}_Z)\Delta^{(\ell+1)}_X\Delta^{(\ell+1)}_Z}}.$
Similarly the negative part gives
\begin{align}
    &\mathcal Z_-=\int_{-\infty}^0 \exp\left[-\frac{1}{\Delta^{(\ell+1)}_X} X_\alpha^{(\ell+1)\mu}Z_\alpha^{(\ell+1)\mu}-\frac{1}{2\Delta^{(\ell+1)}_X}(Z_\alpha^{(\ell+1)\mu})^2+\right.\\&\left.+\frac{1}{\Delta^{(\ell+1)}_Z}Z_\alpha^{(\ell+1)\mu} W^{(\ell)T}_\alpha X^{(\ell)\mu}-\frac{1}{2\Delta^{(\ell+1)}_Z}(Z_\alpha^{(\ell+1)\mu})^2\right]=\sqrt{\frac{\pi}{2a}}e^{q_-^2}\text{erfc}(q_-)
\end{align}
We have $p_-=\frac{1}{1+\mathcal{Z}_+/\mathcal{Z}_-}$,with
\begin{equation}
  \frac{\mathcal{Z}_+}{\mathcal{Z}_-}=\exp\left[\frac{2X_\alpha^{(\ell+1)\mu} W^{(\ell)T}_\alpha X^{(\ell)\mu}}{\Delta^{(\ell+1)}_X+\Delta^{(\ell+1)}_Z}\right] \frac{1+\text{erf}(q_+)}{\text{erfc}(q_-)} 
\end{equation}
The sampling procedure is the same as the sign and the ReLU activation: one first draws a Bernoulli$(p_-)$ variable and then samples either from the negative or positive truncated normal in \eqref{eq:sample_Z_abs}, respectively if the Bernoulli variable is one or zero.
\subsection{Multinomial probit for multiclass classification}
In the setting of multiclass classification we one hot encode the output label $y^\mu\in\{1,2,\dots,C\}$, with $C$ the number of classes. To lighten the notation, in this section we use $Z^{(L+1)}\to Z$\footnote{$Z^{(L+1)}$ here are the preactivations of the last layer.}, $X^{(L)}\to X$, $\Delta_Z^{(L+1)}\mapsto \Delta_Z$, $W^{(L)}\to W $. Hence we have $Z=XW^T+\epsilon_{Z}$, with $\epsilon_{Z}$ a matrix with i.i.d. elements $\mathcal N(0,\Delta_Z)$. We then define the output of the network to be $y^\mu=\argmax_{\alpha\in[C]}Z^{\mu}_\alpha$. This model is also known as the multinomial probit model, and the sampling scheme we describe has been introduced in \cite{albert1993bayesian} and \cite{held2006bayesian}.
We can then do Gibbs sampling over $W$ as usual. We are left with sampling $Z$. Its conditional distribution is
\begin{align}
    &P(Z_\mu|y^\mu,W, X^{\mu})=\frac{P(y^\mu|Z^{\mu})P(Z^{\mu}|W,X)}{P(y^\mu|W,X)}=\\&=\frac{1}{\mathcal Z}\exp\left[-\frac{1}{2\Delta^{(Z)}}\left(Z^{\mu}_\alpha -W^{T}_\alpha X^{\mu}\right)^2\right]\prod_{\alpha\neq y}\Theta(Z^{\mu}_{y^\mu}>Z^{\mu}_\alpha)
\end{align}
Let $Z^\mu_{-\alpha}=(Z^\mu_1,\dots,Z^\mu_{\alpha-1},Z^\mu_{\alpha+1},\dots,Z^\mu_{C})$, i.e., the vector with coordinate $\alpha$ removed.
We then sample $Z^{\mu}$ in  coordinate wise manner. We have 
\begin{align}
    \label{eq:sample_Z_multinomial_probit_notlabel}
    &P(Z^{\mu}_{y^\mu}|y^\mu,W, X^{\mu}, Z^{\mu}_{-y^\mu})\propto \exp\left[-\frac{1}{2\Delta_Z}\left(Z^{\mu}_{y^\mu} -W^{T}_{y^\mu} X^{\mu}\right)^2\right]\mathbb{I}\left[Z^{\mu}_{y^\mu}\geq \max_{\alpha\neq y^\mu}Z^{\mu}_\alpha \right]\\
    & P(Z^{\mu}_\alpha|y^\mu,W, X^{\mu}, Z^{\mu}_{-\alpha})\propto \exp\left[-\frac{1}{2\Delta_Z}\left(Z^{\mu}_\alpha -W^{T}_\alpha X^{\mu}\right)^2\right]\mathbb{I}\left[Z^{\mu}_\alpha\leq Z^{\mu}_{y^\mu}\right],\qquad \forall\alpha\neq y^\mu
    \label{eq:sample_Z_multinomial_probit_label}
\end{align}
These distributions are truncated Gaussians so they are easy to sample. To be more precise on goes through $\alpha=1,2,\dots,C$ sequentially and draws $Z^{\mu}_\alpha$ from \eqref{eq:sample_Z_multinomial_probit_label} or \eqref{eq:sample_Z_multinomial_probit_notlabel}.
\section{Adding biases}
\label{app:biases}
We consider a layer of the kind $Z^{(\ell+1)\mu}_\alpha=W^{(\ell)^T}X^{(\ell)\mu}+b^{(\ell)}+\epsilon_Z^{(\ell+1)}$, with $\epsilon_Z^{(\ell+1)}$  a matrix with i.i.d. $\mathcal{N}(0,\Delta^{(\ell+1)}_Z)$ elements.
We suppose that the biases have a prior $b^{(\ell)}_\alpha\sim\mathcal N(0,1/\lambda^{(\ell)}_b)$.
We now compute the conditional probabilities in this setting. The strategy is to use the expressions obtained before and absorb $b$ in the previous terms.
In the case of $P(W^{(\ell)}_\alpha|X^{(\ell)},Z^{(\ell+1)}_\alpha)$ and $P(X^{(\ell)\mu}|Z^{(\ell)\mu}W^{(\ell)},Z^{(\ell+1)\mu})$, one can replace $Z^{(\ell+1)\mu}_\alpha\to Z^{(\ell+1)\mu}_\alpha- b^{(\ell)}_\alpha$ to obtain the additional conditioning on $b$.
Moving to $P(Z^{(\ell+1)\mu}_\alpha|X^{(\ell+1)\mu}_\alpha,W^{(\ell)}_\alpha, X^{(\ell)\mu})$ one can instead do the substitution $W^{(\ell)T}_\alpha X^{(\ell)\mu}\to W^{(\ell)T}_\alpha X^{(\ell)\mu}+b^{(\ell)}_\alpha$.
We are left with computing $P(b^{(\ell)}_\alpha|X^{(\ell)},Z^{(\ell+1)}_\alpha,W^{(\ell)}_\alpha)$. We have
\begin{align}
&P(b^{(\ell)}_\alpha|X^{(\ell)},Z^{(\ell+1)}_\alpha,W^{(\ell)}_\alpha)\propto\\&\propto\exp\left[-\frac{1}{2\Delta_Z^{(\ell+1)}}\sum_\mu \left(Z^{(\ell+1)\mu}_\alpha-W^{(\ell)T}_\alpha X^{(\ell)\mu}-b^{(\ell)}_\alpha\right)^2-\frac{\lambda^{(\ell)}_b}{2}(b_\alpha^{(\ell)})^2\right]=\\&=\exp\left[-\frac{1}{2}(b^{(\ell)}_\alpha)^2\left(\lambda^{(\ell)}_b+\frac{n}{\Delta_Z^{(\ell+1)}}\right)+\frac{1}{\Delta_Z^{(\ell+1)}}b^{(\ell)}_\alpha\left(\sum_\mu Z^{(\ell+1)\mu}_\alpha-W^{(\ell)T}_\alpha X^{(\ell)\mu}\right)\right]\propto \\&\propto\mathcal{N}(b^{(\ell)}_\alpha|(m^{(\ell)}_b)_\alpha,(\sigma_b^{(\ell)})^2)
\end{align}
with $(\sigma_b^{(\ell)})^2=\frac{\Delta_Z^{(\ell+1)}}{n+\Delta_Z^{(\ell+1)}\lambda^{(\ell)}_b}$ and $(m^{(\ell)}_b)_\alpha=\frac{1}{\Delta_Z^{(\ell+1)}\lambda^{(\ell)}_b+n}\left(\sum_\mu Z^{(\ell+1)\mu}_\alpha-W^{(\ell)T}_\alpha X^{(\ell)\mu}\right).$\\
For the other variables we have
\begin{align}
&P(W^{(\ell)}_\alpha|X^{(\ell)},Z^{(\ell+1)}_\alpha,b^{(\ell)}_\alpha)=\mathcal N(W^{(\ell)}_\alpha|m^{(W)}_\alpha,\Sigma^{(W)}),\\
    &P(Z^{(\ell+1)\mu}_\alpha|X^{(\ell+1)\mu}_\alpha,W^{(\ell)}_\alpha, X^{(\ell)\mu},b^{(\ell)}_\alpha)=\exp\left[-\frac{1}{2\Delta_Z^{(\ell+1)}}\left(Z^{(\ell+1)\mu}_\alpha-W^{(\ell)T}_\alpha X^{(\ell)\mu}-b^{(\ell)}_\alpha\right)^2\right.\\&-\left.\frac{1}{2\Delta_X^{(\ell+1)}}\left(\sigma(Z^{(\ell+1)\mu}_\alpha)-X^{(\ell+1)\mu}_\alpha\right)^2\right],\\
    &P(X^{(\ell)\mu}|Z^{(\ell)\mu},W^{(\ell)},Z^{(\ell+1)\mu, b^{(\ell)}})=\mathcal N\left(X^{(\ell)\mu}|m^{(X_\ell)\mu},\Sigma^{(X_\ell)}\right)
\end{align}
with 
\begin{align}
    &\Sigma^{(\ell)}_W=\left(\frac{1}{\Delta_Z^{(\ell+1)}} X^{(\ell)T}X^{(\ell)}+\lambda^{(\ell)}_W\mathbb I_{d}\right)^{-1},\quad (m^{(\ell)}_W)_\alpha=\frac{1}{\Delta_Z^{(\ell+1)}}\Sigma^{(W)} X^{(\ell)T}\left(Z^{(\ell+1)}_\alpha- \boldsymbol{1}_n b^{(\ell)}_\alpha\right)\\
    & \Sigma^{(\ell)}_X=\left(\frac{1}{\Delta_Z^{(\ell+1)}}W^{(\ell)T}W^{(\ell)}+\frac{1}{\Delta^{(\ell)}_X}\mathbb I_{d_\ell}\right)^{-1},\\&(m^{(\ell)}_X)^\mu=\Sigma^{(X_\ell)}\left(\frac{1}{\Delta^{(\ell)}_X}\sigma(Z^{(\ell)\mu})+\frac{1}{\Delta_Z^{(\ell+1)}}W^{(\ell)T}(Z^{(\ell+1)\mu}-b^{(\ell)})\right).
\end{align}
Here $\boldsymbol{1}_n$ is the vector of length $n$ whose coordinates are all ones.

There is an alternative way of sampling $b$. One can consider $b^{(\ell)}$ as the $d_\ell+1$th column of an extended weight matrix $\Tilde W^{(\ell)}=[W^{(\ell)}|b^{(\ell)}]$. These extended weights act on $\Tilde X^{(\ell)}=\left[X^{(\ell)}|\boldsymbol{1}_n\right]$, a $n\times (d_\ell+1)$ matrix, whose last column contains all ones. The generative process, with biases included, can be written as $Z^{(\ell+1)}=\Tilde{X} \Tilde{W}^T+\epsilon_Z^{(\ell+1)}$, with $\epsilon_Z$ an $n\times d_{\ell+1}$ matrix with i.i.d. $\mathcal N(0,\Delta_Z^{(\ell+1)})$ elements. One can compute (and later sample from) $P(\Tilde W^{(\ell)}|\Tilde{X}^{(\ell)},Z^{(\ell+1)})$, following the same procedure that we used to compute \eqref{eq:cond_W}. The only difference is that the prior variance is not uniform over $\Tilde W$: one will instead have both $\lambda_W^{(\ell)}$ and $\lambda_b^{(\ell)}$.
\section{Gibbs sampling for convolutional neural networks}
\label{app:cnn}
In order to implement the Gibbs sampler for convolutional networks, we need to sample from the conditional distribution of variables involved in convolutional layers and pooling layers.
\subsection{Convolutional Layer}
In this appendix we we formulate the Gibbs sampler in the case of convolutional layers. The difficulty in this case stems from the structure in the weights. In this section we use the following notation
\begin{itemize}
    \item $C_\ell$ is the number of channels in layer $\ell$
    \item $H_W^{(\ell)},W_W^{(\ell)}$ is the height and width of the convolutional filter $W^{(\ell)}$.
    \item $H_X^{(\ell)},W_X^{(\ell)}$ are the height and width of the input $X^{(\ell)}$.
    \item $H_Z^{(\ell+1)},W_X^{(\ell+1)}$ are the height and width of the output $Z^{(\ell+1)}$.
    \item $K_\ell$ is the size (i.e. $K_\ell=H_W^{(\ell)} \times W_W^{(\ell)}$) of each channel of filter $W^{(\ell)}$.
    \item $d_\ell$, is the sizes of each channel in layer $\ell$(i.e., the total number of variables at layer $\ell$ is $C_\ell\times d_\ell$)
    \item $\alpha,\alpha'\in[C_{\ell+1}]$ are the indices for the channel in layer $\ell+1$
    \item $\beta,\beta'\in[C_\ell]$ are the indices for channels in layer $\ell$.
    \item $a,a'\in[d_{\ell+1}]$ are indices for positions inside layer $\ell+1$ (i.e. $a$ specifies both the horizontal and vertical position within the layer)
    \item $b,b',c,c'\in[d_\ell]$ are indices for positions in layer $\ell$.
    \item $r,r'\in[K_\ell]$ are indices for the position within the filter (e.g. if the filter is $3\times 3$ then $K=9$ and $r$ runs over all the components of the filter)
    \item $i=(\beta,r)$ and $i'=(\beta',r')$ are used to group pairs of indices.
\end{itemize}
Commas will be used to separate indices, whenever there is ambiguity.
The basic building block is the noisy convolutional layer, which has the following expression
\begin{equation}
\label{eq:cnn_layer_no_bias}
    Z^{(\ell+1)\mu}_{\alpha a}=\sum_{\beta=1}^{C_\ell}\sum_{r=1}^{K_\ell} \,W^{(\ell)}_{\alpha\beta r}\;X^{(\ell)\mu}_{\beta, \nu_a(r)}\;+\;(\epsilon^{(\ell+1)}_Z)^\mu_{\alpha a},
\end{equation}
with $(\epsilon^{(\ell+1)}_Z)^\mu_{\alpha a}\sim\mathcal N(0,\Delta_Z^{(\ell+1)})$.
$W\in\mathbb{R}^{C_{\ell+1}\times C_\ell \times K_\ell}$.
We also indicate with $\nu_a(r)$ the position of the $r-th$ coordinate of the filter inside the input layer, when the output is in position $a$. In other words $\nu_a:[K_\ell]\mapsto [d_\ell]$.
First we notice that $P(Z^{(\ell+1)}|W^{(\ell)},X^{(\ell)},X^{(\ell+1)})$ is basically unaffected by the structure of weights. We will concentrate on computing $P(W^{(\ell)}|Z^{(\ell+1)},X^{(\ell)})$ and $P(X^{(\ell)}|Z^{(\ell+1)},Z^{(\ell)},W^{(\ell)})$.
Let us begin by
\begin{align}
    &P(W^{(\ell)}_\alpha|X^{(\ell)},Z^{(\ell+1)}_\alpha)=\\&=\exp\left[-\frac{1}{2\Delta_Z^{(\ell+1)}}\sum_{a,\mu}\left(Z^{(\ell+1)\mu}_{\alpha a}-\sum_{\beta,r}W^{(\ell)}_{\alpha\beta r}\;X^{(\ell)\mu}_{\beta,\nu_a(r)}\right)^2-\frac{\lambda^{(\ell)}_W}{2}\sum_{\beta,r}(W^{(\ell)}_{\alpha\beta r})^2\right]=\\&
    =\exp\left[-\frac{1}{2}\sum_{\beta,r,\beta',r'}W^{(\ell)}_{\alpha\beta r}\left(\frac{1}{\Delta_Z^{(\ell+1)}}\sum_{\mu,a}X^{(\ell)\mu}_{\beta,\nu_a(r)}X^{(\ell)\mu}_{\beta',\nu_a(r')}+\lambda^{(\ell)}_W\delta_{rr'}\delta_{\beta\beta'}\right)W^{(\ell)}_{\alpha\beta' r'}\right.\\&\left.+\frac{1}{\Delta_Z^{(\ell+1)}}\sum_{\beta,r}W^{(\ell)}_{\alpha\beta r}\sum_{a,\mu}Z^{(\ell+1)\mu}_{\alpha a}X^{(\ell)\mu}_{\beta,\nu_a(r)}\right] =\mathcal{N}(W^{(\ell)}_\alpha|(m^{(\ell)}_W)_{\alpha},\Sigma^{(\ell)}_W)
\end{align}
Computing these quantities requires grouping the two indices $r,\beta$ into a single index and then inverting the matrix of the quadratic form in $W$. 
The double index matrix we would like to invert is 
\begin{equation}
    \Tilde{A}_{\beta r, \beta' r'}=\frac{1}{\Delta_Z^{(\ell+1)}}\sum_{\mu,a}X^{(\ell)\mu}_{\beta,\nu_a(r)}X^{(\ell)\mu}_{\beta',\nu_a(r')}+\lambda^{(\ell)}_W\delta_{rr'}\delta_{\beta\beta'}
\end{equation}
So we define $i=(\beta,r)$.
Let $A_{ii'}=\Tilde{A}_{\beta(i) r(i), \beta'(i') r'(i')}=A\left(\frac{1}{\Delta_Z^{(\ell+1)}}\sum_{\mu,a}X^{(\ell)\mu}_{\beta(i),\nu_a(r(i))}X^{(\ell)\mu}_{\beta'(i'),\nu_a(r'(i'))}+\lambda^{(\ell)}_W\delta_{r(i),r'(i')}\delta_{\beta(i),\beta'(i')}\right)$
then we define $\Sigma^{(W)}=A^{-1}$. Finally we define $\widetilde{A^{-1}}_{\beta r\beta' r'}=A^{-1}_{i(\beta,r),i'(\beta',r')}$. In words, we're packing the indices of $\tilde A$, inverting it, and then unpacking the indices. For the mean we have
\begin{equation}
    (m^{(\ell)}_W)_{\alpha \beta r}=\frac{1}{\Delta_Z^{(\ell+1)}} \sum_{a,\mu} Z^{(\ell+1)\mu}_{\alpha a}\sum_{\beta', r'}[\widetilde{A^{-1}}]_{\beta r\beta' r'} \;X^{(\ell)\mu}_{\beta',\nu_a(r')}
\end{equation}
We now look at $P(X^{(\ell)}|Z^{(\ell+1)},Z^{(\ell)},W^{(\ell)})$.
In the computation we use the identity $\sum_{c}\delta_{c,\nu_a(b)}=1$
\begin{align}
\label{eq:P_X_giv_W_Z_conv2d}
&P(X^{(\ell)\mu}|Z^{(\ell+1)\mu},Z^{(\ell)\mu},W^{(\ell)})=\\&=\exp\left[-\frac{1}{2\Delta_X^{(\ell)}}\sum_{\beta,b}\left(X^{(\ell)\mu}_{\beta b}-\sigma(Z^{(\ell)\mu}_{\beta, b})\right)^2-\frac{1}{2\Delta_Z^{(\ell+1)}}\sum_{a,\alpha}\left(Z^{(\ell+1)\mu}_{\alpha a}-\sum_{\beta,r}W^{(\ell)}_{\alpha\beta r}X^{(\ell)\mu}_{\beta \nu_a(r)}\right)^2\right]\\&
=\exp\left[-\frac{1}{2\Delta_X^{(\ell)}}\sum_{\beta, b}\left(X^{(\ell)\mu}_{\beta b}-\sigma(Z^{(\ell)\mu}_{\beta b})\right)^2\right.\\&\left.-\frac{1}{2\Delta_Z^{(\ell+1)}}\sum_{a,\alpha}\left(Z^{(\ell+1)\mu}_{\alpha a}-\sum_{\beta,r}W^{(\ell)}_{\alpha\beta r}\sum_{c}\delta_{c,\nu_a(r)}X^{(\ell)\mu}_{\beta \nu_a(r)}\right)^2\right]\propto\\&
\exp\left[-\frac{1}{2}\sum_{c,\beta,c',\beta'}X^{(\ell)\mu}_{\beta c}\left(\frac{1}{\Delta_X^{(\ell)}}\delta_{cc'}\delta_{\beta\beta'}+\frac{1}{\Delta_Z^{(\ell+1)}}\sum_{r,r',a,\alpha}W^{(\ell)}_{\alpha\beta r}\;W^{(\ell)}_{\alpha\beta' r'}\;\delta_{c,\nu_a(r)}\delta_{c',\nu_a(r')}\right)X^{(\ell)\mu}_{\beta' c'}\right.\\&\left.+\sum_{c,\beta}X^{(\ell)\mu}_{\beta c}\left(\frac{1}{\Delta_Z^{(\ell+1)}}\sum_{a,\alpha,r}\delta_{c,\nu_a(r)}Z^{(\ell+1)\mu}_{\alpha a} W^{(\ell)}_{\alpha \beta r}+\frac{1}{\Delta^{(X)}_{\ell}}\sigma(Z^{(\ell)\mu}_{\beta c})\right)\right]=\\&=\mathcal N(X^{(\ell)\mu}|(m^{(\ell)}_X)^\mu,\Sigma^{(\ell)}_X)
\end{align}
As in the previous case we let $A$ be the matrix of the quadratic form, with 
\begin{align}
    &\Tilde A_{\beta c \beta' c'}=\frac{1}{\Delta_X^{(\ell)}}\delta_{cc'}\delta_{\beta\beta'}+\frac{1}{\Delta_Z^{(\ell+1)}}\sum_{r,r',a,\alpha}W^{(\ell)}_{\alpha\beta r}\;W^{(\ell)}_{\alpha\beta' r'}\;\delta_{c,\nu_a(r)}\delta_{c',\nu_a(r')}=\\&\frac{1}{\Delta_X^{(\ell)}}\delta_{cc'}\delta_{\beta\beta'}+\frac{1}{\Delta_Z^{(\ell+1)}}\sum_{\alpha}\sum_{\substack{a:\\c, c'\in\nu_a([K_\ell])}}\sum_{r,r'}W^{(\ell)}_{\alpha\beta, \nu^{-1}_a(r)}\;W^{(\ell)}_{\alpha\beta', \nu^{-1}_a(r')}
\end{align}
In the last passage $\nu_a([K_\ell])$ indicates the image of the whole filter through $\nu_a$. We basically incorporated the constraint that $a$ should be such that $c,c'$ are in the same subset of the input (if such $a$s exist at all).
As we did previously for $W$ we group the indices using $i=(\beta, c)$ and $i'=(\beta',c')$.
We then define the matrix $A_{ii'}=\Tilde{A}_{\beta(i)c(i)\beta'(i')c'(i')}$, giving $\Sigma^{(\ell)}_X=A^{-1}$. 
For the mean we have
\begin{align}
    &(m^{(\ell)}_X)^\mu_{\beta c}=\sum_{c',\beta'}[\widetilde {A^{-1}}]_{\beta c \beta' c'}\left(\frac{1}{\Delta_Z^{(\ell+1)}}\sum_{a,\alpha,r}\delta_{c',\nu_a(r)}\;Z^{(\ell+1)\mu}_{\alpha a} W^{(\ell)}_{\alpha \beta' r}+\frac{1}{\Delta^{(X)}_{\ell}}\sigma(Z^{(\ell)\mu}_{\beta' c'})\right)=\\&=\frac{1}{\Delta^{(\ell)}_X}\sum_{c',\beta'}[\widetilde{ A^{-1}}]_{\beta c \beta' c'}\sigma(Z^{(\ell)\mu}_{\beta' c'})+\frac{1}{\Delta_Z^{(\ell+1)}}\sum_{a,\alpha,r,\beta'}[\widetilde {A^{-1}}]_{\beta c \beta' \nu_a(r)} Z^{(\ell+1)\mu}_{\alpha a} W^{(\ell)}_{\alpha \beta' r}
\end{align}
\subsection{Practical implementation}
So far we packed the spatial (i.e. x and y coordinate within an image) in a single index. This allowed for more agile computations. We now unpack the indices and translate the results we obtained.
In a practical case $X^{(\ell)}\in\mathbb{R}^{n\times C_\ell\times H_X^{(\ell)}\times W_X^{(\ell)}}$. For the weights we have $W^{(\ell)}\in\mathbb{R}^{C_{\ell+1}\times C_\ell\times H_W^{(\ell)}\times W_W^{(\ell)}}$.
Let $s_y,s_x$ be respectively the strides along the $y,x$ axes. We do not use any padding. Then the height and width of $Z^{(\ell+1)}$ will be respectively $H_Z^{(\ell+1)}=\lfloor \frac{H_X^{(\ell)}-H_W^{(\ell)}}{s_y}\rfloor+1$ and $W_Z^{(\ell+1)}=\lfloor \frac{W_X^{(\ell)}-W_W^{(\ell)}}{s_x}\rfloor+1$.

Given $a=(a_y,a_x)$ the position inside layer $\ell+1$ one can write $\nu_a(r)=\nu_a(r_y,r_x)=(r_y+s_y a_y, r_x+s_x a_x)$, yielding the following expression for the forward pass:
\begin{equation}
    Z^{(\ell+1)\mu}_{\alpha a_y a_x}=\sum_{\beta=0}^{C_\ell-1}\sum_{r_x=0}^{W_W^{(\ell)}-1}\sum_{r_y=0}^{H_W^{(\ell)}-1} \,W^{(\ell)}_{\alpha\beta r_y r_x}\;X^{(\ell)\mu}_{\beta, r_y+s_y a_y,r_x+s_x a_x}\;+\;\epsilon^{(Z_{(\ell+1)})}
\end{equation}
In $\epsilon$ we removed all the indices, for notation simplicity.\\
Recall we hat to compute the matrix $\Tilde{A}$ when sampling $W$.
In this notation the expression for $\Tilde{A}\in\mathbb{R}^{C_\ell\times H_W^{(\ell)}\times W_W^{(\ell)}\times C_\ell\times H_W^{(\ell)}\times W_W^{(\ell)}}$ becomes
\begin{align}
    \Tilde{A}_{\beta r_y r_x \beta' r'_y r'_x}&=\frac{1}{\Delta_Z^{(\ell+1)}} \sum_{\mu=0}^{n-1}\sum_{a_y=0}^{H_Z^{(\ell+1)}-1}\sum_{a_x=0}^{W_Z^{(\ell+1)}-1} X^{(\ell)\mu}_{\beta, r_y+s_y a_y,r_x+s_x a_x} X^{(\ell)\mu}_{\beta', r_y'+s_y a_y,r_x'+s_x a_x}+\\&+\lambda_W^{(\ell)}\delta_{r_y, r'_y}\delta_{r_x, r'_x}\delta_{\beta, \beta'}
\end{align}

The expression for $m_W$ in turn becomes
\begin{align}
    &(m^{(\ell)}_W)_{\alpha\beta r_yr_x}=\\&\frac{1}{\Delta_Z^{(\ell+1)}} \sum_{\mu=0}^{n-1}\sum_{a_y=0}^{H_Z^{(\ell+1)}-1} \sum_{a_x=0}^{W_Z^{(\ell+1)}-1} Z^{(\ell+1)\mu}_{\alpha a_y a_x}\sum_{\beta'=0}^{C_\ell-1}\sum_{ r'_y=0}^{H_W^{(\ell)}-1}\sum_{ r'_x=0}^{W_W^{(\ell)}-1}[\widetilde{ A^{-1}}]_{\beta r_y r_x, \beta' r'_y r'_x} \;X^{(\ell)\mu}_{\beta', r_y'+s_y a_y,r_x'+s_x a_x}
\end{align}
We now move to sampling $X^{(\ell)\mu}$. In that case we have 
\begin{equation}
    \Tilde A_{\beta c \beta' c'}=\frac{1}{\Delta_X^{(\ell)}}\delta_{cc'}\delta_{\beta\beta'}+\frac{1}{\Delta_Z^{(\ell+1)}}\sum_{r,r',a,\alpha}W^{(\ell)}_{\alpha\beta r}\;W^{(\ell)}_{\alpha\beta' r'}\;\delta_{c,\nu_a(r)}\delta_{c',\nu_a(r')}
\end{equation}
\begin{align}
    &\Tilde A_{\beta c_yc_x \beta' c'_y c'_x}=\frac{1}{\Delta_X^{(\ell)}}\delta_{c_yc'_y}\delta_{c_xc'_x}\delta_{\beta\beta'}+\\&+\frac{1}{\Delta_Z^{(\ell+1)}}\sum_{r_y,r_x,r'_yr'_x,a_y,a_x,\alpha}W^{(\ell)}_{\alpha\beta r_y r_x}\;W^{(\ell)}_{\alpha\beta' r'_y r'_x}\;\delta_{c_y,r_y+s_y a_y}\delta_{c_x,r_x+s_x a_x}\delta_{c'_y,r'_y+s_ya_y}\delta_{c'_x,r'_x+s_xa_x}
\end{align}
For $m^{(\ell)}_X$ we have
\begin{align}
    &(m^{(\ell)}_X)_{\beta c_x c_y}=\frac{1}{\Delta_X^{(\ell)}}\sum_{\beta', c'_y,c'_x} [\widetilde{A^{-1}}]_{\beta c_y c_x \beta' c'_y c'_x} \sigma(Z^{(\ell)\mu}_{\beta' c'_y c'_x})+\\&+\frac{1}{\Delta_Z^{(\ell+1)}} \sum_{\alpha,a_y,a_x,r_y,r_x,\beta'}[\widetilde{A^{-1}}]_{\beta c_y c_x \beta', r_y+s_ya_y,r_x+s_x a_x}Z^{(\ell+1)\mu}_{\alpha a_y a_x}W^{(\ell)}_{\alpha\beta' r_y r_x}
\end{align}
One can see that when the filter $W^{(\ell)}$ has dimensions (height and width) that are much smaller than those of $X^{(\ell)}$, then  $\Tilde{A}$ will have few nonzero elements. In fact for $\Tilde A_{\beta c_y c_x \beta' c'_y c'_x}$ to be nonzero, one must have that $c,c'$ are close enough to be contained in the filter $W$. This implies that all the pairs of pixels $c,c'$ with $|c_x-c'_x|>W_W^{(\ell)}$ or $|c_y-c'_y|>H_W^{(\ell)}$ will have $\Tilde A_{\beta c_y c_x \beta' c'_y c'_x}=0$. Hence $\tilde{A}$ will be a sparse tensor. The same will somewhat be true in the covariance, which is the inverse of $A$.
\subsection{Average Pooling}
Here we look at how to put pooling into the mix. We focus on average pooling, which is easier since it is a linear transformation. 
For each pixel $b\in [d_\ell]$, let $P_\ell(b)$\footnote{notice that for shape mismatch issues, some pixels could be mapped into nothing, so $P$ only acts on the pixels that get pooled} be the "pooled pixel" in layer $\ell+1$ to which $b$ gets mapped. Hence we have $P:[d_\ell]\mapsto [d_{\ell+1}]$, a surjective function.
Given a pixel $a$ in layer $\ell+1$ this will have multiple preimages through $P$, we denote the set of preimages as $P^{-1}(a)$. $P^{-1}(a)$ can therefore be seen as the receptive field of pixel $a$.
Let
\begin{equation}
    X^{(\ell+1)\mu}_{\beta a}=\frac{1}{|P^{-1}(a)|}\sum_{b\in P^{-1}(a)}X^{(\ell)\mu}_{\beta b}+(\epsilon^{(\ell+1)}_X)^\mu_{\beta a}
\end{equation}
be the generative model for the pooling layer. Notice we inject some noise $(\epsilon^{(\ell+1)}_X)^\mu_{\beta a}\sim\mathcal N(0,\Delta_X^{(\ell+1)})$ in the output.
The probability of $X^{(\ell)}$ factorizes according to the pooling receptive fields, i.e., pixels in different receptive fields are independent.
Suppose also that we have $X^{(\ell)\mu}_{\beta b}=\sigma(X^{(\ell-1)\mu}_{\beta b})+(\epsilon^{(\ell)}_X)^\mu_{\beta b}$, with $(\epsilon^{(\ell)}_X)^\mu_{\beta b}\sim\mathcal N(0,\Delta_X^{(\ell)})$. $\sigma(\cdot)$ here can be an element wise activation function, but it can also represent any other transformation \footnote{for example the same scheme can be used to have a convolutional layer followed by a pooling layer. Just set $\sigma(x)=x$ and $X^{(\ell)}=\text{Conv2d}(W^{(\ell-2)},X^{(\ell-2)})$ (notice the absence of noise). Here Conv2d executes the 2d convolution between its inputs.}.
We have
\begin{align}
&P(\{X^{(\ell)\mu}_{\beta,b}\}_{b\in P^{-1}(a)}|\{X^{(\ell-1)\mu}_{\beta b}\}_{b\in P^{-1}(a)}, X^{(\ell+1)\mu}_{\beta,a})=P(X^{(\ell+1)\mu}_{\beta a}|\{X^{(\ell)\mu}_{\beta,b}\}_{b\in P^{-1}(a)})\times\\&\times\prod_{b\in P^{-1}(a)}P(X^{(\ell)\mu}_{\beta b}|X^{(\ell-1)\mu}_{\beta b})\propto\exp\left[-\frac{1}{2\Delta^{(\ell)}_X}\sum_{b\in P^{-1}(a)}\left(X^{(\ell)\mu}_{\beta b}-\sigma(X^{(\ell-1)\mu}_{\beta b})\right)^2\right.\\&\left.-\frac{1}{2\Delta_X^{(\ell+1)}}\left(X^{(\ell+1)\mu}_{\beta a}-\frac{1}{|P^{-1}(a)|}\sum_{b\in P^{-1}(a)}X^{(\ell)\mu}_{\beta b}\right)
^2\right]\propto \\&\exp\left[-\frac{1}{2}\sum_{b,c\in P^{-1}(a)}X^{(\ell)\mu}_{\beta b}\left(\frac{1}{\Delta_X^{(\ell)}}\delta_{bc}+\frac{1}{\Delta_X^{(\ell+1)}|P^{-1}(a)|^2}\right)X^{(\ell)\mu}_{\beta c}+\right.\\&\left.+\sum_{b\in P^{-1}(a)} X^{(\ell)\mu}_{\beta b}\left(\frac{\sigma(X^{(\ell-1)\mu}_{\beta b})}{\Delta_X^{(\ell)}}+\frac{X^{(\ell+1)\mu}_{\beta a}}{\Delta_X^{(\ell+1)}|P^{-1}(a)|}\right)\right]\propto \mathcal{N}(\{X^{(\ell)\mu}_{\beta,b}\}_{b\in P^{-1}(a)}|m^{(\ell)}_X)^\mu_\beta,\Sigma^{(\ell)}_X)
\end{align}

With $\Sigma^{(\ell)}_X \in \mathbb{R}^{P^{-1}(a)\times P^{-1}(a)}$\footnote{In principle the size depends on $a$, however normally each pixel in layer $\ell+1$ has the same number of preimages.} as $\Sigma^{(\ell)}_X=\left(\frac{1}{\Delta_X^{(\ell)}}\mathbb{I}+\frac{1}{\Delta_X^{(\ell+1)}|P^{-1}(a)|^2}\boldsymbol 1\boldsymbol{1}^T\right)^{-1} $ and
$(m^{(\ell)}_X)^\mu_{\beta b}=\sum_b(\Sigma^{(\ell)}_X)_{bc}\left(\frac{\sigma(X^{(\ell-1)\mu}_{\beta c})}{\Delta_X^{(\ell)}}+\frac{X^{(\ell+1)\mu}_{\beta a}}{\Delta_X^{(\ell+1)}|P^{-1}(a)|}\right)$.
Exploiting the fact that $\Sigma^{(\ell)}_X$ is a projector plus the identity one can simplify the previous expressions. We use the fact that for a matrix $A=r\mathbb I+s vv^T$, with $r,s\in\mathbb{R},\; v\in \mathbb R^n$, its inverse is $A^{-1}=\frac{1}{r}\mathbb I-\frac{s}{r^2+rs||v||^2} v v^T$. In out present case $r=1/\Delta_X^{(\ell)},\; s=\frac{1}{\Delta_X^{(\ell+1)}|P^{-1}(a)|^2}$ and $v=\boldsymbol{1}$, $v\in\mathbb{R}^{|P^{-1}(a)|}$.
This gives (keeping implicit that $a=P(b)$),
\begin{align} 
     &(m^{(\ell)}_X)^\mu_{\beta b}=\sigma(X^{(\ell-1)\mu}_{\beta b})+\frac{\Delta_X^{(\ell)}}{\Delta_X^{(\ell)}+|P^{-1}(a)|\Delta_X^{(\ell+1)}}\left(X^{(\ell+1)\mu}_{\beta a}-\frac{1}{|P^{-1}(a)|}\sum_{c\in P^{-1}(a)}\sigma(X^{(\ell-1)\mu}_{\beta c})\right)\\
    &\Sigma^{(\ell)}_X=\E\left[X^{(\ell)\mu}_\beta (X^{(\ell)\mu }_\beta)^T \right]-(m^{(\ell)}_X)^\mu_{\beta}\,(m^{(\ell)}_X)^{\mu T}_{\beta}=\\&=\Delta_X^{(\ell)} \mathbb{I}-\frac{(\Delta_X^{(\ell)})^2}{|P^{-1}(a)|(|P^{-1}(a)|\Delta_X^{(\ell+1)}+\Delta_X^{(\ell)})}\boldsymbol{1}\boldsymbol 1^T\nonumber
\end{align}
To generate a Gaussian variable with this covariance we employ the following trick. Suppose $\Sigma= r \mathbb I_d -s vv^T$ (i.e., identity minus a projector) and $z\sim \mathcal{N}(0,r\mathbb{I}_d)$. Define $\Bar{z}=z-q \langle z,v\rangle$, with $q=\frac{1}{||v||^2}\left(1-\sqrt{1-\frac{s||v||^2}{r}}\right)$. Then $\Bar{z}\sim\mathcal{N}(0,\Sigma)$. In our case, we have $v=\boldsymbol{1}$, $r=\Delta_X^{(\ell)}$, $s=\frac{(\Delta_X^{(\ell)})^2}{|P^{-1}(a)|(|P^{-1}(a)|\Delta_X^{(\ell+1)}+\Delta_X^{(\ell)})}$. This gives $q=\frac{1}{|P^{-1}(a)|}\left(1-\sqrt{\frac{|P^{-1}(a)|\Delta_X^{(\ell+1)}}{|P^{-1}(a)|\Delta_X^{(\ell+1)}+\Delta^{(\ell)}_X}}\right)$

It can happen that the pooling layer size is not perfectly matched to the image size (i.e., $H^{(\ell)}_X/H_W^{(\ell)}$ is not an integer). In this case, we define $H_X^{(\ell+1)}=\lfloor H^{(\ell)}_X/H_W^{(\ell)} \rfloor$ (basically we discard the last part of the input layer). The pixels that do not contribute to $X^{(\ell+1)}$ should be sampled from $\mathcal{N}(\sigma(X^{(\ell-1)\mu}_{\beta b}),\Delta_X^{(\ell)})$.

Sampling $X^{(\ell+1)}$ does not require any additional custom function. In fact
$P(X^{(\ell+1)}|W^{(\ell+1)},X^{(\ell+2)},X^{(\ell)})$ is equal to \eqref{eq:P_X_giv_W_Z_conv2d}, where $\Delta_X^{(\ell)}\mapsto \Delta_X^{(\ell+1)}$, $\Delta_Z^{(\ell+1)}\mapsto \Delta^{(X)}_{\ell+2}$, and the nonlinearity $\sigma$ is replaced by the pooling layer expression.

\subsection{Biases}
In the convolutional networks the biases are introduced by writing the layer as
\begin{equation}
\label{eq:cnn_layer_with_bias}
    Z^{(\ell+1)\mu}_{\alpha a}=\sum_{\beta=1}^{C_\ell}\sum_{r=1}^{K_\ell} \,W^{(\ell)}_{\alpha\beta r}\;X^{(\ell)\mu}_{\beta, \nu_a(r)}\;+b^{(\ell)}_\alpha+\;(\epsilon_Z^{(\ell+1)})^\mu_{\alpha a},
\end{equation} 
where $b^{(\ell)}_\alpha\sim\mathcal{N}(0,1/\lambda^{(\ell)}_b)$, and $(\epsilon_Z^{(\ell+1)})^\mu_{\alpha a}\sim\mathcal N(0,\Delta_Z^{(\ell+1)})$.
Notice that there is one bias parameter per channel.
Sampling $W^{(\ell)}, X^{(\ell)}$ is very similar to the previous case.
One must simply replace $Z^{(\ell+1)\mu}_{\alpha a}\mapsto Z^{(\ell+1)\mu}_{\alpha a}-b^{(\ell)\alpha}$.
When sampling $Z^{(\ell+1)\mu}_{\alpha a}$ instead one should instead replace $\sum_{\beta r}W^{(\ell)}_{\alpha\beta r}X^{(\ell)\mu}_{\beta\nu_a(r)}\mapsto \sum_{\beta r}W^{(\ell)}_{\alpha\beta r}X^{(\ell)\mu}_{\beta\nu_a(r)}-b^{(\ell)}_\alpha$.

The update equation for the biases is
\begin{align}
&P(b^{(\ell)}_\alpha|Z^{(\ell+1)}_\alpha,X^{(\ell)},W^{(\ell)})\propto P(Z^{(\ell+1)}_\alpha|b^{(\ell)}_\alpha,X^{(\ell)},W^{(\ell)})P(b^{(\ell)}_\alpha|X^{(\ell)},W^{(\ell)})=\\&=P(Z^{(\ell+1)}_\alpha|b^{(\ell)}_\alpha,X^{(\ell)},W^{(\ell)})P(b^{(\ell)}_\alpha)\propto\\&\propto
\exp\left[-\frac{1}{2}\left(\frac{n d_{\ell+1}}{\Delta_Z^{(\ell+1)}}+\lambda^{(\ell)}_b\right)(b^{(\ell)}_\alpha)^2+\frac{1}{\Delta_Z^{(\ell+1)}}\sum_{\mu a}\left(Z^{(\ell+1)\mu}_{\alpha a}-\sum_{\beta r}W^{(\ell)}_{\alpha\beta r}X^{(\ell)\mu}_{\beta\nu_a(r)}\right)\right]=\\&=\mathcal{N}\left(b^{(\ell)}_\alpha\middle|\frac{\sum_{\mu a}\left(Z^{(\ell+1)\mu}_{\alpha a}-\sum_{\beta r}W^{(\ell)}_{\alpha\beta r}X^{(\ell)\mu}_{\beta\nu_a(r)}\right)}{n d_{\ell+1}+\Delta_Z^{(\ell+1)}\lambda^{(\ell)}_b},\frac{\Delta_Z^{(\ell+1)}}{n d_{\ell+1}+\lambda^{(\ell)}_b\Delta_Z^{(\ell+1)}}\right)
\end{align}

\section{Other Monte Carlo algorithms}
\label{app:other_mcmc}
In this appendix, we provide the pseudocode for HMC on the classical posterior, HMC on the intermediate noise posterior and MALA on the classical posterior
\subsection{Hamiltonian Monte Carlo}
\label{app:HMC}
We provide the pseudocode of HMC (algorithm \ref{algo:HMC}). In the paper we used the implementation from \cite{TensorflowProbability} for synthetic data and the one in \cite{cobb2020scaling} for real-world data. The algorithm depends on the learning rate and the number of leapfrog steps, which are hyperparameters that need to be optimized appropriately. 
\begin{algorithm}[H]
\begin{algorithmic}
\State{\textbf{Input:} 
Probability measure to sample $\pi(x)$, initial condition $x_0$, step size $\eta$, number of leapfrog steps $L$, length of the simulation $t_\text{max}$.}
\State\textbf{Output: }a sequence $S$ of samples
\State $S\gets [x_0]$
\For{$i=1$ to $t_\text{max}$}
    \State{$p \sim \mathcal{N}(0,\mathbb{1})$} \Comment{Sample a new momentum from a normal distribution}
    \State{$x \gets x_0$}
    \State{$\mathcal{H} \gets \pi(x_0) - \frac{\|p\|^2}{2}$}
    \For{$j=1$ to $L$} \Comment{Simulate Hamiltonian dynamics to propose a new state}
        \State{$p \gets p - \frac{\epsilon}{2} \nabla \pi(x)$} \Comment{Simulate a half-step for momentum}
        \State{$x \gets x + \epsilon p$} \Comment{Simulate a full-step for position}
        \State{$p \gets p - \frac{\epsilon}{2} \nabla \pi(x)$} \Comment{Simulate another half-step for momentum}
    \EndFor
    \State{$\mathcal{H}_{\text{prop}} \gets \pi(x) - \frac{\|p\|^2}{2}$} \Comment{Compute the proposed Hamiltonian}
    \State $z\sim \rm{Uniform}([0,1])$
    \If{$\log(z) < \mathcal{H} - \mathcal{H}_{\text{prop}}$} \Comment{Accept/Reject the proposal}
        \State{$x_0 \gets x$}
    \EndIf
    \State $S\text{.append}(x_0)$
\EndFor
\end{algorithmic}
\caption{Hamiltonian Monte Carlo}
\label{algo:HMC}
\end{algorithm}

\subsection{Metropolis Adjusted Langevin Algorithm}
\label{app:MALA}
The pseudocode for MALA is reported here (algorithm \ref{alg:MALA}). In the paper we used the implementation in \cite{TensorflowProbability} for synthetic data and a custom implementation for real data. The algorithm has on one hyper-parameter: the learning rate. If the learning rate is too large the Metropolis acceptance rate gets too low. Lowering the learning rate increases the acceptance rate up to an optimal value, after which the acceptance rate starts decreasing again.

\begin{algorithm}[H]
\caption{Metropolis Adjusted Langevin Algorithm (MALA)}
\label{alg:MALA}
\begin{algorithmic}
\State\textbf{Input: } Probability measure to sample $\pi(\cdot)$, initial condition $x$, step size $\eta$, length of the simulation $t_\text{max}$.
\State\textbf{Output: }a sequence $S$ of samples
\State $S\gets [x]$
\For{$t=1,\dots,t_\text{max}$}
    \State $x'\gets x+\eta\nabla\pi(x)+\sqrt{2\eta}\mathcal N(0,1)$ \Comment{The Gaussian noise is i.i.d. over the coordinates of $W$}
    \State $P_{x\to x'}\gets \exp\left[-\frac{1}{4\eta}(x'-x-\eta\nabla\pi(x))^2\right]$
    \State $P_{x'\to x}\gets \exp\left[-\frac{1}{4\eta}(x-x'-\eta\nabla\pi(x'))^2\right]$
    \State $P_\text{acc}\gets\min\left\{1, \frac{\pi(x')P_{x'\to x}}{\pi(x)P_{x\to x'}}\right\}$\Comment{Acceptance probability according to Metropolis rule}
    \State $z\sim\text{Uniform}([0,1])$
    \If{$z<P_\text{acc}$} 
    \State $x'\gets x$ \Comment{If accepted update $x$}
    \EndIf
    \State $S\text{.append}(x)$
\EndFor
\end{algorithmic}
\end{algorithm}

\section{Synthetic data experiments}
\label{app:synth_data_exp}
This appendix provides additional details about the numerical experiments on synthetic data presented in section \ref{sec:numerics_gibbs}. We recall that in these experiments we ran Gibbs on the intermediate noise posterior, MALA on the classical posterior and HMC both on the intermediate and classical posterior. For MALA and HMCs we used the implementation contained in \cite{TensorflowProbability}. To produce the left plot in figure \ref{fig:therm_synth}, we ran each algorithm with 3 logarithmically spaced values of $\Delta$ per decade, both from the informed and uninformed initialization. For each value of $\Delta$ and each initialization we further ran each algorithm 72 times, each time with different teacher network, random student initialization (whenever this initialization was used) and noise in the MCMC. For the Gibbs sampler we set all variables to zero in the uninformed initialization. Instead for the HMCs and MALA, still in the uninformed case, we found that initializing the variables as i.i.d. Gaussians with standard deviation $10^{-4}$ helped the algorithms thermalize. We remark that not having to tune the initialization norm, is another advantage of the Gibbs sampler. 

The HMCs and MALA all have hyperparameters to select. The optimal parameters were obtained by doing a grid search on the learning rate and, in the case of HMC, number of leapfrog steps. 
For each value of $\Delta$ we re-optimize the hyperparameters. In the case of HMC, we build our grid by trying 3 learning rates per decade and two numbers of leapfrog steps per decade. In the case of MALA we try three learning rates per decade. For both MALA and HMC we select the hyperparameters for which the test MSE takes the least time to descend and subsequently become stationary. The optimal values of the hyper parameters, as well as the number of steps in each simulation, are in tables \ref{tab:HMC_classical}, \ref{tab:HMC_intermediate} for HMC and tables \ref{tab:MALA_info}, \ref{tab:MALA_zero} for MALA. For Gibbs all experiments were run for $2.5\times 10^6$ steps. The number of steps of the different algorithms has been chosen so that the runs executed within 5:30h. All experiments were run on one core of Intel Xeon Platinum 8360Y running at 2.4 GHz.
\begin{table}[htbp]
    \centering
    \caption{Parameters for HMC with intermediate noise posterior on syntetic data. HMC Steps is the total number of Metropolis steps during the run. For the plots in Figure \ref{fig:therm_synth} a measurement is taken every 10 steps.}
    \label{tab:HMC_intermediate}
    \begin{tabular}{cccc}
        \hline
        Delta & Learning Rate & Leapfrog Steps & HMC Steps \\
        \hline
        $1.0 \times 10^0$ & $5.0 \times 10^{-4}$ & $10^3$ & $10^4$ \\
        $4.64 \times 10^{-1}$ & $5.0 \times 10^{-4}$ & $10^3$ & $10^4$ \\
        $2.15 \times 10^{-1}$ & $5.0 \times 10^{-4}$ & $10^3$ & $10^4$ \\
        $1.0 \times 10^{-1}$ & $5.0 \times 10^{-4}$ & $10^3$ & $10^4$ \\
        $4.64 \times 10^{-2}$ & $5.0 \times 10^{-4}$ & $10^3$ & $10^4$ \\
        $2.15 \times 10^{-2}$ & $5.0 \times 10^{-4}$ & $10^3$ & $10^4$ \\
        $1.0 \times 10^{-2}$ & $5.0 \times 10^{-4}$ & $10^3$ & $10^4$ \\
        $4.64 \times 10^{-3}$ & $5.0 \times 10^{-5}$ & $10^3$ & $10^5$ \\
        $2.15 \times 10^{-3}$ & $5.0 \times 10^{-5}$ & $10^3$ & $10^5$ \\
        $1.0 \times 10^{-3}$ & $5.0 \times 10^{-5}$ & $10^3$ & $10^5$ \\
        $4.64 \times 10^{-4}$ & $5.0 \times 10^{-5}$ & $10^3$ & $10^5$ \\
        $2.15 \times 10^{-4}$ & $5.0 \times 10^{-5}$ & $10^3$ & $10^5$ \\
        $1.0 \times 10^{-4}$ & $5.0 \times 10^{-5}$ & $10^3$ & $10^5$ \\
        $4.64 \times 10^{-5}$ & $5.0 \times 10^{-5}$ & $10^3$ & $10^5$ \\
        $2.15 \times 10^{-5}$ & $5.0 \times 10^{-5}$ & $10^3$ & $10^5$ \\
        $1.0 \times 10^{-5}$ & $5.0 \times 10^{-5}$ & $10^3$ & $10^5$ \\
        \hline
    \end{tabular}
\end{table}
\begin{table}[htbp]
    \centering
    \caption{Parameters for HMC with classical posterior on syntetic data. HMC Steps is the total number of Metropolis steps during the run. For the plots in Figure \ref{fig:therm_synth} a measurement is taken every 10 steps.}
    \label{tab:HMC_classical}
    \begin{tabular}{cccc}
        \hline
        Delta & Learning Rate & Leapfrog Steps & HMC Steps \\
        \hline
        $1.0 \times 10^0$ & $5.0 \times 10^{-4}$ & $2.0 \times 10^1$ & $10^4$ \\
        $4.64 \times 10^{-1}$ & $5.0 \times 10^{-4}$ & $2.0 \times 10^1$ & $10^4$ \\
        $2.15 \times 10^{-1}$ & $5.0 \times 10^{-4}$ & $2.0 \times 10^1$ & $10^4$ \\
        $1.0 \times 10^{-1}$ & $5.0 \times 10^{-4}$ & $2.0 \times 10^1$ & $10^4$ \\
        $4.64 \times 10^{-2}$ & $5.0 \times 10^{-4}$ & $2.0 \times 10^1$ & $10^4$ \\
        $2.15 \times 10^{-2}$ & $5.0 \times 10^{-5}$ & $10^2$ & $10^5$ \\
        $1.0 \times 10^{-2}$ & $5.0 \times 10^{-5}$ & $10^2$ & $10^5$ \\
        $4.64 \times 10^{-3}$ & $5.0 \times 10^{-5}$ & $10^3$ & $10^5$ \\
        $2.15 \times 10^{-3}$ & $5.0 \times 10^{-5}$ & $10^3$ & $10^5$ \\
        $1.0 \times 10^{-3}$ & $5.0 \times 10^{-5}$ & $10^3$ & $10^5$ \\
        $4.64 \times 10^{-4}$ & $5.0 \times 10^{-5}$ & $10^3$ & $10^5$ \\
        $2.15 \times 10^{-4}$ & $5.0 \times 10^{-5}$ & $10^3$ & $10^5$ \\
        $1.0 \times 10^{-4}$ & $5.0 \times 10^{-5}$ & $10^3$ & $10^5$ \\
        \hline
    \end{tabular}
\end{table}

\begin{table}[htbp]
    \centering
    \caption{Parameters for MALA on synthetic data with informed initialisation. Langevin Steps is the number of steps during each run. Measurements of the test MSE, as reported in Figure \ref{fig:therm_synth} are taken every \textit{Spacing} steps.}
    \label{tab:MALA_info}
    \begin{tabular}{cccc}
        \hline
        Delta & Learning Rate & Langevin Steps & Spacing \\
        \hline
        $1.0 \times 10^0$ & $1.0 \times 10^{-5}$ & $10^5$ & $10^2$ \\
        $4.64 \times 10^{-1}$ & $1.0 \times 10^{-5}$ & $10^5$ & $10^2$ \\
        $2.15 \times 10^{-1}$ & $1.0 \times 10^{-5}$ & $10^5$ & $10^2$ \\
        $1.0 \times 10^{-1}$ & $1.0 \times 10^{-5}$ & $10^5$ & $10^2$ \\
        $4.64 \times 10^{-2}$ & $1.0 \times 10^{-5}$ & $10^6$ & $10^2$ \\
        $2.15 \times 10^{-2}$ & $1.0 \times 10^{-6}$ & $10^6$ & $10^2$ \\
        $1.0 \times 10^{-2}$ & $1.0 \times 10^{-6}$ & $10^6$ & $10^2$ \\
        $4.64 \times 10^{-3}$ & $1.0 \times 10^{-6}$ & $10^6$ & $10^2$ \\
        $2.15 \times 10^{-3}$ & $1.0 \times 10^{-6}$ & $10^6$ & $10^2$ \\
        $1.0 \times 10^{-3}$ & $1.0 \times 10^{-6}$ & $10^6$ & $10^2$ \\
        $4.64 \times 10^{-3}$ & $1.0 \times 10^{-7}$ & $10^6$ & $10^2$ \\
        $2.15 \times 10^{-3}$ & $1.0 \times 10^{-7}$ & $10^6$ & $10^2$ \\
        $1.0 \times 10^{-4}$ & $1.0 \times 10^{-7}$ & $10^6$ & $10^2$ \\
        $4.64 \times 10^{-4}$ & $1.0 \times 10^{-9}$ & $1.1\times 10^7$ & $1.1 \times 10^{3}$ \\
        $2.15 \times 10^{-4}$ & $1.0 \times 10^{-9}$ & $1.1\times 10^7$ & $1.1 \times 10^{3}$ \\
        $1.0 \times 10^{-5}$ & $1.0 \times 10^{-9}$ & $1.1\times 10^7$ & $1.1 \times 10^{3}$ \\
        \hline
    \end{tabular}
\end{table}

\begin{table}[htbp]
    \centering
    \caption{Parameters for MALA on synthetic data with zero initialisation. Langevin Steps is the number of steps during each run. Measurements of the test MSE, as reported in Figure \ref{fig:therm_synth} are taken every \textit{Spacing} steps.}
    \label{tab:MALA_zero}
    \begin{tabular}{cccc}
        \hline
        Delta & Learning Rate & Langevin Steps & Spacing \\
        \hline
        $1.0 \times 10^0$ & $1.0 \times 10^{-5}$ & $10^5$ & $10^2$ \\
        $4.64 \times 10^{-1}$ & $1.0 \times 10^{-5}$ & $10^5$ & $10^2$ \\
        $2.15 \times 10^{-1}$ & $1.0 \times 10^{-5}$ & $10^6$ & $10^2$ \\
        $1.0 \times 10^{-1}$ & $1.0 \times 10^{-5}$ & $10^6$ & $10^2$ \\
        $4.64 \times 10^{-2}$ & $1.0 \times 10^{-5}$ & $10^6$ & $10^2$ \\
        $2.15 \times 10^{-2}$ & $1.0 \times 10^{-6}$ & $10^6$ & $10^2$ \\
        $1.0 \times 10^{-2}$ & $1.0 \times 10^{-7}$ & $1.1 \times10^7$ & $1.1 \times 10^{3}$ \\
        $4.64 \times 10^{-3}$ & $1.0 \times 10^{-7}$ & $1.1 \times10^7$ & $1.1 \times 10^{3}$ \\
        $2.15 \times 10^{-3}$ & $1.0 \times 10^{-7}$ & $1.1 \times10^7$ & $1.1 \times 10^{3}$ \\
        $1.0 \times 10^{-3}$ & $1.0 \times 10^{-8}$ & $1.1 \times10^7$ & $1.1 \times 10^{3}$ \\
        $4.64 \times 10^{-3}$ & $1.0 \times 10^{-8}$ & $1.1 \times10^7$ & $1.1 \times 10^{3}$ \\
        $2.15 \times 10^{-3}$ & $1.0 \times 10^{-9}$ & $1.1 \times10^7$ & $1.1 \times 10^{3}$ \\
        $1.0 \times 10^{-4}$ & $1.0 \times 10^{-9}$ & $1.1 \times10^7$ & $1.1 \times 10^{3}$ \\
        $4.64 \times 10^{-4}$ & $1.0 \times 10^{-10}$ & $1.1 \times10^7$ & $1.1 \times 10^{3}$ \\
        $2.15 \times 10^{-4}$ & $1.0 \times 10^{-10}$ & $1.1 \times10^7$ & $1.1 \times 10^{3}$ \\
        $1.0 \times 10^{-5}$ & $1.0 \times 10^{-11}$ & $1.1 \times10^7$ & $1.1 \times 10^{3}$ \\
        $4.64 \times 10^{-6}$ & $1.0 \times 10^{-11}$ & $1.1 \times10^7$ & $1.1 \times 10^{3}$ \\
        \hline
    \end{tabular}
\end{table}
\section{Real data experiments}
\label{app:real_data_experiments}
In this appendix, we provide the details of the experiments conducted in sections \ref{sec:real_data_experiments}.
We start by describing the architectures used.
In the case of the MLP with 12 hidden units, the model is
\begin{align}
    \label{eq:argmax_mlp_2_layer}
    f_\text{MLP}(x,W)=\argmax_{\alpha}\left[ W^{(2)T}_\alpha\sigma(W^{(1)}x+ b^{(1)})+b^{(2)}_\alpha \right],
\end{align}
with $x\in\mathbb R^{784}, \; W^{(1)}\in\mathbb R^{12\times 784}, \; b^{(1)}\in\mathbb R^{12}, \; W^{(2)}\in\mathbb R^{10\times 12},\; b^{(2)}\in\mathbb R^{10}$ and $\sigma(x)=\max(0,x)$. $\alpha$ is the row index of $W^{(2)}$.
To apply the Gibbs sampler we must translate this architecture into a posterior using the intermediate noise model. To do so, the additional variables $Z^{(2)}\in\mathbb R^{n\times 12},\;X^{(2)}\in\mathbb R^{n\times 12}, Z^{(3)}\in\mathbb R^{n\times 10}$ are introduced, with $n=6\times 10^4$ in the case of MNIST. A noise with variance $\Delta$ is put on $Z^{(2)},\;X^{(2)}$. Notice there is no noise between $Z^{(3)}$ and the labels $y$. Hence the posterior has the hard constraint $y^\mu=\arg\max_{\alpha\in\{0,1,\dots,9\}} Z^{(3)\mu}_\alpha$. 
The priors on the parameters are given by: $\lambda_W^{(1)}=\lambda_b^{(1)}=784, \; \lambda_W^{(2)}=\lambda_b^{(2)}=12$.
On the intermediate noise posterior we ran experiments with $\Delta=2$, and all variables were set to zero at initial condition.

In the case of the classical posterior, we ran experiments with $\Delta=2$ using MALA and HMC as algorithms. The value of $\Delta$ was picked so that the test error at stationarity is the same as in the intermediate noise posterior. The optimal parameters of HMC are a learning rate of $10^{-3}$ and $200$ leapfrog steps, while for MALA the optimal learning rate is $2\times 10^{-6}$. For HMC, MALA the variables were initialized as i.i.d. Gaussians with respective standard deviations $10^{-1},10^{-4}$.

\begin{figure}
    \centering
    \includegraphics{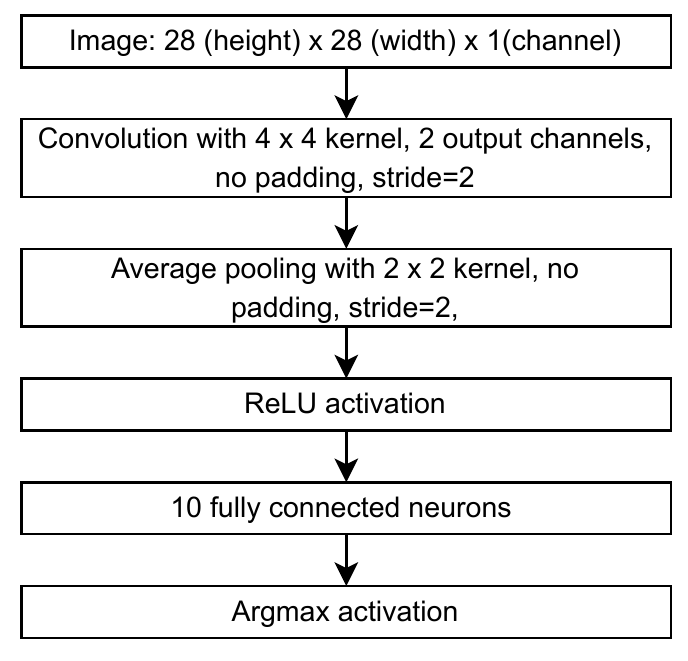}
    \caption{CNN architecture used in the experiments of section \ref{sec:real_data_experiments}. The convolutional layer is composed of the  filter $W^{(1)}$ with shape $2\times1\times4\times4$ and a output channel bias $b^{(1)}\in\mathbb R^2$. The final layer instead has weights $W^{(2)}\in\mathbb R^{72\times 10}$ and bias $b^{(2)}\in \mathbb R^{10}$.}
    \label{fig:cnn_architecture}
\end{figure}

Regarding the CNN, we provide a schematic representation of the architecture in figure \ref{fig:cnn_architecture}. 
In the intermediate noise model a noise is added after the convolution, after the average pooling, after the ReLU, and after the fully connected layer. All noises are i.i.d. Gaussians with variance $\Delta$, as prescribed by the intermediate noise model. To complete the description, we specify the prior. We set $\lambda_W^{(1)}=\lambda_b^{(1)}=16,\;\lambda_W^{(2)}=\lambda_b^{(2)}=72$.

In the intermediate noise posterior, we run the Gibbs sampler with $\Delta=100$, and initialize all variables to zero. For the classical posterior, we run MALA and HMC on the CNN architecture, with $\Delta=10$. This value of $\Delta$ leads to approximately the same
test error as in the intermediate noise posterior.
For HMC we use a learning rate of $10^{-3}$ and $50$ leapfrog steps, while for MALA  we choose $5\times 10^{-6}$ as learning rate. 

Both for MLP and CNN, in the case of the classical posterior, we have to specify a loss function. To do so we replace the argmax in the last layer by a softmax and apply a cross entropy loss on top of the softmax. Calling $Q\in\mathbb R^{n\times 10}$ the output of the softmax, the loss function is $\ell(y^\mu,Q^\mu)=-\log Q^\mu_{y^\mu}$. The argmax is however still used when making predictions, for example when evaluating the model on the test set. All experiments were run on one NVIDIA V100 PCIe 32 GB GPU and one core of Xeon-Gold running at 2.1 GHz.

\end{document}